%% file: acl2022.tex
\pdfoutput=1

\documentclass[11pt]{article}

\usepackage{authblk}
\usepackage{acl}

\usepackage{times}
\usepackage{latexsym}

\usepackage[T1]{fontenc}

\usepackage[utf8]{inputenc}

\usepackage{microtype}

\usepackage{graphicx}
\usepackage{amsmath, amssymb}
\usepackage{makecell}
\usepackage{multirow}
\usepackage{rotating}
\usepackage{bigstrut}
\usepackage{tabularx}
\usepackage{longtable}
\usepackage{xcolor}
\usepackage{todonotes}
\usepackage{ltxtable}
\usepackage{booktabs}
\usepackage{float}
\usepackage{caption}
\usepackage{subcaption}

\title{Achieving Reliable Human Assessment of Open-Domain Dialogue Systems}

\author[1,3]{\textbf{Tianbo Ji}}
\author[1,4]{\textbf{Yvette Graham}}
\author[1,3]{\textbf{Gareth Jones}}
\author[2,3]{\textbf{Chenyang Lyu}}
\author[5]{\textbf{Qun Liu}}
\affil[1]{ADAPT Centre}
\affil[2]{SFI Centre for Research Training in Machine Learning}
\affil[3]{School of Computing, Dublin City University}
\affil[4]{School of Computer Science and Statistics, Trinity College Dublin}
\affil[5]{Noah's Ark Lab, Huawei}

\affil[ ]{
\texttt{\{tianbo.ji,gareth.jones\}@adaptcentre.ie}, \texttt{graham.yvette@gmail.com},
\protect\\ \texttt{chenyang.lyu2@mail.dcu.ie}, \texttt{qun.liu@huawei.com}
}

\begin{document}
\maketitle
\input{sec-00-abstract}
\input{sec-01-intro}
\input{sec-02-relwork}

\input{sec-03-method}
\input{sec-04-eval}

\input{sec-05-exp-persona}
\input{sec-06-exp-topics}

\input{sec-07-auto-metrics}

\input{sec-07-conc}

\newpage

\bibliography{anthology,custom}
\bibliographystyle{acl_natbib}

\input{sec-08-appendix}

\end{document}

%% file: sec-00-abstract.tex
\begin{abstract}
  Evaluation of open-domain dialogue systems is highly challenging and development of better techniques is highlighted time and again as desperately needed.
  Despite substantial efforts to carry out  reliable live evaluation of systems in recent competitions, annotations have been abandoned and reported as too unreliable to yield sensible results. This is a serious problem since automatic metrics are not known to provide a good indication of what may or may not be a high-quality conversation.
  Answering the distress call of competitions that have emphasized the urgent need for better evaluation techniques in dialogue, we present the successful development of human evaluation that is  highly reliable while still remaining feasible and low cost. Self-replication experiments reveal almost perfectly repeatable results with a correlation of $r=0.969$.
  Furthermore, due to the lack of appropriate methods of statistical significance testing, the likelihood of potential improvements to systems occurring due to chance is rarely taken into account in dialogue evaluation, and the evaluation we propose facilitates application of standard tests.
  Since we have developed a highly reliable evaluation method, new insights into system performance can be revealed. We therefore include a comparison of state-of-the-art models (i) with and without personas, to measure the contribution of personas to conversation quality, as well as (ii) prescribed versus freely chosen topics. Interestingly with respect to personas, results indicate that personas do not positively contribute to conversation quality as expected.

\end{abstract}

%% file: sec-01-intro.tex
\section{Introduction}

Evaluation of open-domain dialogue is particularly challenging and has been cited in high-profile competitions as a known open problem \cite{convai2paper}. 
Challenges arise primarily from the fact that in real-world conversations there exists such a vast number of possible appropriate responses.
Subsequently, dialogue evaluation that relies on comparison with pre-created reference dialogues  incur substantial false-negative rates as many  appropriate responses are unfairly penalized simply for not corresponding closely with references.
In addition, evaluation faces further challenges with respect to the ability to fully take into account dialogue history.\footnote{The protocol employed in this work was approved by the DCU Research Ethics Committee.}

In this paper, we present a new method of open-domain dialogue evaluation based on human assessment of live conversations with models that
avoids the need for pre-created reference dialogues 
and ensures full familiarity with dialogue history, ticking two important boxes in terms of validity. 
Although live human evaluation of models has the advantage of being highly valid, 
reliability unfortunately cannot be assumed and  developing methods of evaluation for language tasks that achieve high rater consistency has been challenging, often resulting in low levels of agreement between annotators \cite{current-evaluation-dialogue,wmt11,wmt12,wmt13,wmt14,metric-usr}.
Despite challenges in this respect, our proposed method provides highly reliable evaluation, achieving a correlation of $r=0.969$ in self-replication experiments. 
Additionally, the evaluation can be carried out cheaply and on a large scale through strict quality controlled crowd-sourcing, as well as including score standardization for fairer ranking of competing models. We make the data and code  publicly available to aid future research.\footnote{\url{https://github.com/TianboJi/Dialogue-Eval}}

%% file: sec-02-relwork.tex
\section{Problems in Past Evaluations}



A common issue occurs that can potentially impact the validity of results is \emph{filtering the set of systems to be evaluated via automatic metric scores}. Since metric scores are known to be a poor substitute for human assessment, this only results in the possibility that the best system according to human judges is inadvertently filtered out at this stage.
For example, ConvAI2 \cite{convai2paper} ranked models firstly using automatic metrics before top models according to metric scores were assessed by crowd-sourced workers on Mechanical Turk, while similarly in the sixth Dialog System Technology Challenge (DSTC6) systems were  filtered according to metric scores prior to human evaluation.

In terms of the live evaluation, competitions such as Convai2 report such evaluations as highly challenging, with many of the resulting dialogues  reported to be \emph{senseless,  offensive},  or  \emph{simply  not  in  line  with  instructions} and ultimately  live evaluation results have been discarded. 

Despite challenges, competitions that operate in the public domain, making data and evaluation techniques available to researchers (such as ourselves) should be applauded for such efforts. 

On the other hand, competitions that (for one reason or another) do not release data and evaluation techniques into the public domain have reported relative success in terms of human evaluation. However until such methods can be accessed and independently verified through replication studies, they will unfortunately have little impact .
The first Amazon Alexa Socialbot Grand Challenge required human assessors to score how \emph{coherent} and \emph{engaging}  conversations were on a 1--5 rating scale by two distinct groups: volunteer Amazon employees (experts), and general Alexa users (crowds) \cite{alexa1}, are reported to achieve a correlation of overall scores for the two types of human assessors at $0.93$. The absolute average rating across all chatbots was reported to be 20\% lower for experts compared to general users.
In an additional effort to evaluate models, \emph{conversational user experience, coherence, engagement, domain coverage, topical diversity}, and \emph{conversational depth} were assessed (1--5 scale), with combined scores reported to correlate with those of general users at $r=0.66$.
In addition to methods and data not being publicly available, correlations are difficult to interpret since no detail is provided about the number of judgments on which the correlation is calculated for example.

In addition to competitions that generally aim to include human evaluation of systems, automatic metrics are often proposed for dialogue evaluation, themselves requiring a human evaluation data set on which to evaluate the proposed metric.
However, inappropriate statistics are often applied.
For example, ~\newcite{metric-towards-holistic} propose a holistic metric to automatically evaluate four distinct aspects of dialogue,
and a human evaluation experiment is deployed on Mechanical Turk using a 1--5 rating scale. 
The mean correlation between human assessors is reported as $r=0.61$. However, mean correlations are unfortunately difficult to interpret, since correlation coefficients are not additive , averages calculated in the usual way cannot be assumed to reflect central tendency, and unfortunately, the distribution of correlations is not reported \cite{alexander1990note}.

\newcite{metric-usr} propose USR (UnSupervised and Reference-free), an unsupervised model that predicts the quality of dialog for a range of criteria using various rating scales:
\emph{understandable} (0--1 rating scale), \emph{natural} (1--3), \emph{maintains context} (1--3), \emph{interesting} (1--3), \emph{uses knowledge} (0--1); \emph{overall quality} (1--5).
Despite human evaluation being carried out by experts 
inter-annotator agreement levels varied depending on criteria being measured, ranging from as low as $0.298$.
Additionally, although correlations between human assessments are reported as significant at $p<0.01$, despite such statistics often being reported for correlations, they are unfortunately not very  meaningful in terms of their impact on correlation interpretation and can be somewhat misleading. Contrary to common expectations, even small effect sizes (low $r$) can produce very low p-values (strong significance) in such tests.
Aiming to achieve a \emph{significant correlation} is an extremely low bar to reach in terms of consistency, since a low p-value in this case simply rejects the null hypothesis that \emph{the correlation is zero}.

In addition to the above issues, human evaluation of dialogue systems rarely take into account the fact that differences in performance can occur simply by chance. The method of human evaluation we propose provides a means of applying standard tests for statistical significance to avoid concluding differences that are highly likely to have occurred simply by chance.

%% file: sec-03-method.tex
\section{Crowd-sourcing Reliable Human Assessment of Open-Domain Dialogue}

Crowd-sourcing with highly accurate quality control provides a potential mechanism to ensure the three most important criteria that makes an evaluation meaningful while still remaining feasible: validity, reliability and scalability.
Subsequently, we ask crowd-workers to carry out live text-based chat with models prior to that same worker also rating the quality of the immediately preceding conversation. 

\subsection{Human Ratings of Dialogue Quality}

A continuous (0--100) rating scale is employed with three main motivation points 
\cite{graham-etal-2013-crowd,novikova-etal-2018-rankme, li2019acute, santhanam-shaikh-2019-towards, santhanam2020studying, mille-etal-2020-third,wmt20,howcroft-etal-2020-twenty_years}. 
Firstly, continuous scales reduce potential bias when comparing the performance of competing models by enabling \emph{score standardization}. The score distribution of each human assessor is standardized according to overall mean and standard deviation of all ratings provided by that assessor, thus removing any adverse effects of those employing overly harsh (or indeed lenient) scoring strategies.
Secondly, the 0--100 rating scale allows 
\emph{standard significance tests} to score distributions of models to help determine which models significantly outperform others.
Thirdly, and possibly most importantly, a continuous rating scale facilitates \emph{highly accurate quality control} of crowd-sourced workers so that the evaluation can scale while still maintaining validity at a low cost.

Each human assessor is firstly asked to carry out a live conversation with a randomly selected model, comprised of a minimum of 10 conversational inputs, before rating the quality of the conversation that just took place under a number of criteria shown in 
Figure \ref{fig:criteria}. Note that the measurement criteria we employed are not immutable and we encourage to extend or adjust the criteria for future studies as necessary.
\begin{figure}[htpb]
    \centering
    \small
    \input{tables/tbl-02-evaluation-label}
    \caption{Criteria employed to assess models in our human evaluation in the form of Likert statements; corresponding evaluation labels (left) \emph{not} shown to human assessors.}
    \label{fig:criteria}
\end{figure}

A continuous rating scale is advantageous for several reasons but employment of such a scale raises the question of how it should be labeled.
In evaluation of language tasks, adjectival scale labels, such as \emph{poor, low, medium, high, perfect/ okay, good, excellent}, and so on, are often employed despite their likely contribution to annotator \emph{inconsistency} \cite{loukina-etal-2020-using, sorodoc-etal-2017-multimodal}.
This is despite evidence of adjectival scale labels being problematic in terms of bias resulting from positively and negatively worded items not being true opposites of one another, and items intended to have neutral intensity in fact proving to have specific conceptual meanings. \newcite{alexandrov2010characteristics} provides a summary of  issues associated with adjectival labels.

To avoid any such causes of inconsistency, we structure each rating as a simple Likert declarative statement and ask human assessors to rate the degree to which they agree with each of these statement, making it possible to keep the rating scale constant while only changing the statement for each measurement criteria.
We ask judges to rate each conversation under the seven aforementioned measurement criteria (Figure \ref{fig:criteria})
along with a continuous rating scale labeled only at each extreme with \emph{strongly disagree} (left); \emph{strongly agree} (right). 



\subsection{Quality Controlling the Crowd for Open-Domain Dialogue}

We structure Human Intelligence Tasks (HITs) so
that a sufficiently rich score distribution is collected from each individual worker who participated, asking each to hold six conversations, comprised of a shuffled arrangement of five dialogue models and a single quality control model.

Many approaches to quality controlling the crowd employ  gold-standard items as quality checks~\cite{gold-standard-02,gold-standard-03}.
This approach is however highly likely to allow low quality data to pollute the resulting evaluation, since any worker willing to assign high scores to \emph{all items} will undeservedly pass this check.\footnote{Anecdotally, in our experience many workers on crowd-sourcing platforms attempt this strategy to pass quality checks.} 
The approach also runs in contrast to our aim of the same individual who took part in a live conversation to also assess its quality, as it relies on the use of  \emph{pre-created} gold standard conversations.

Our quality control approach overcomes these challenges by deploying models in live conversations that have \emph{known distinct performance levels} instead of asking workers to assess the quality of pre-existing known high quality conversations.
Within a HIT, the five models $m$ can produce \emph{some quality level of conversation} and the model $l$ produces known \emph{lower} quality dialogues (lower than the five models). For a single worker who takes part in conversations with $m$ and $l$, we then check how consistently the worker rated the conversations of $l$ lower than $m$.
This results in a quality control mechanism that does not ask workers to be consistent with other workers or to correctly rate gold standard dialogues but only assesses worker consistency by how \emph{consistently they distinguish between known distinct performance models and only with respect to their own conversation ratings}.




From a practical standpoint, creating a low performance model, $l$, is additionally far less challenging and costly than pre-creating a known set of \emph{high quality} dialogues, and degraded models operate fully automatically.
Low quality models produce outputs via generation of  \emph{random responses} with \emph{meaning distortion} also applied. 

For \emph{random response} degradation: Low quality responses are generated by random sampling responses from training set dialogues with the intention of disregarding any previous input from the user, so responses from the model are likely to be perceived as low quality since they have low relevance.
To reduce the quality of conversations further, we apply \emph{meaning distortion}: each response, $r$, is altered to distort its meaning by randomly selecting a sequence of words within that response and replacing it with a sequence of words sampled from a distinct  training set dialogue, with the length of the replaced word sequence being determined by the number of words in $r$. The specific details are provided in Appendix \ref{replacement}, and Figure \ref{tbl:bad-conversation} in Appendix  \ref{sec:app-experiments} gives a typical example.


Hits subsequently consist of a total of six dialogues comprised of five genuine models and a single quality control model that generates meaning distorted and random responses. Crowd-sourced workers converse with each model before rating conversation quality (model order is shuffled and blind).
Statistical significance tests are then applied to score distributions of workers for the ratings they  attributed to ordinary models, $m$, relative to the low quality model, $l$.
The resulting $p$-value is then employed as a means of rating worker consistency,
and any worker with $p>=0.05$ shows no significant difference between low and ordinary model quality and is filtered out.


\subsection{Calculating System-Level Scores}

Scores are collected from workers who rate models on a 0--100 rating scale,
and we refer to these scores as \emph{raw} scores.
Scores for negative attributes, i.e. \emph{robotic} and \emph{repetitive}, are then reversed for ease of further comparison, $100$ $-$ the original rating.
A distribution of scores is extracted for each worker
and raw scores are standardized according to each worker's mean and standard deviation, in order to iron out any differences in worker scoring strategy. 

Average standardized scores for each criteria are calculated, and an overall score is calculated as the average of all measurement criteria.

%% file: tables/tbl-02-evaluation-label.tex


\begin{tabularx}{0.48\textwidth}{rX}
\toprule
\emph{Robotic}:   & \emph{It was obvious that I was talking to a chatbot as opposed to another human user.}   \\
\emph{Interesting}:& \emph{The conversation with the chatbot was interesting.}  \\
\emph{Fun}:        & \emph{The conversation with the chatbot was fun/enjoyable.}   \\
\emph{Consistent}: & \emph{The chatbot was consistent throughout the conversation.}  \\
\emph{Fluent}:     & \emph{The chatbot's English was fluent and natural throughout the conversation.}   \\
\emph{Repetitive}: & \emph{I felt that the chatbot kept being repetitive during the conversation.}   \\
\emph{Topic}:      & \emph{The chatbot stays on topic.}   \\
\bottomrule
\end{tabularx}

%% file: sec-04-eval.tex
\section{Meta-Evaluation}\label{meta}

\begin{table*}[htbp]
  \centering
  \small
  \input{tables/tbl-00-data}

  \caption{Numbers of workers who took part in human evaluation of models, average time taken per dialogue in minutes (min), and total number of dialogues assessed before and after quality control in which workers freely chose the topic (Free run 1); precisely the same experiment set-up was repeated (Free run 2); where the topic was prescribed via an ice-breaker statement (Ice-breaker) selected directly from the persona of the model.}
  \label{tbl:passrate}
\end{table*}

In order to assess the reliability of the proposed method of human
evaluation, we carry out a meta-evaluation in which we firstly examine individual human assessor consistency, before conducting a self-replication experiment.
A number of models are required to function as a sample set of test systems, and for this purpose we employ available pre-trained models from ParlAI:\footnote{\url{https://parl.ai/docs/zoo.html}}
Poly-Encoder Transformer~\cite{polyencoder}, Bi-Encoder Transformer~\cite{biencoder2}, Sequence to Sequence~\cite{seq2seq}, Key-Value Memory Networks~\cite{kvmn}, and a  LSTM-based Model~\cite{lstm-based}. Within the evaluation setting of ConvAI2, each model is with a persona consisting of approximately five textual statements to emulate a personality. However, to increase the number of models and to provide an interesting comparison, we additionally include a version of each of the above models without
any persona, resulting in 10 competing models.

Hits 
were posted on the crowd-sourcing platform Amazon Mechanical Turk.\footnote{\url{https://www.mturk.com}}
Firstly, and in order to evaluate the \emph{open-domain} models in as realistic a setting as possible, we allow workers to \emph{choose the topic} of conversation and input their chosen topic in a text field.
The open nature of conversations should be noted however as something that influences the difficulty of producing consistent results in our self-replication experiment.
The fact that we allow human assessors to freely choose the topic of conversation means that differences in ratings could result from legitimate differences in performance when  different topics are chosen by  human assessors. 
We nonetheless test our evaluation allowing the user to choose the topic as this is part of our core aim for developing evaluation of dialogue truly in the open domain. 

Besides choosing a topic, we additionally asked workers to input their opinion of the topic they chose to discuss with models, categorizing the topic as either \emph{liked, ambivalent about it, or disliked}. For example, if the topic they chose to discuss was \emph{dogs}, we were curious to know if this was motivated by the fact that the worker liked or disliked dogs or indeed that they had chosen to discuss something they had no particular feeling about. Table  \ref{tbl:percentage-topic} shows subsequent proportions (\%) of workers, and the detailed instructions are introduced in Figure \ref{instructions} in Appendix \ref{sec:app-experiments}.
Perhaps unsurprisingly, the vast majority of workers chose to discuss something they \emph{liked} (84\% for workers who passed quality control). Nonetheless  7\% of good workers were ambivalent about the topic they chose and  9\% chose a topic they reported as disliking.

\begin{table}[ht]
 \centering
 \input{tables/tbl-17-percentage-topic}
   \caption{Proportions (\%) of topics that are reported as liked, ambivalent about or disliked by workers who passed and failed quality control.}
  \label{tbl:percentage-topic}
\end{table}

Table \ref{tbl:passrate} shows the number of workers who
participated in the initial data collection run  who freely chose the topic of conversation with models (Free run 1), amounting to 1,525 dialogues $\times$ 7 criteria $=$ 10,675 human ratings. The details of payment to each worker and the total experiment cost are provided in Appendix \ref{sec:app-payment}.
Table \ref{tbl:passrate} also shows the proportion of workers who passed quality checks, numbers of dialogues assessed in total before and after quality filtering, as well as the average time taken for workers to complete a hit and average time taken to assess dialogues.
As mentioned previously, we carry out a second data collection
run with precisely the same settings (Free run 2) to measure the reliability of results and Table \ref{tbl:passrate} shows equivalent statistics with respect to Free run 2 in which a total of 1,480 dialogues $\times$ 7 ratings $=$ 10,360 human ratings were collected in total.


\subsection{Human Assessor Consistency}
\label{ind-level}

Although the overall aim of our evaluation is to produce
reliable results at the system level, which we test
later in Section \ref{system-level-test},
 we firstly
examine ratings of workers at the level of individual dialogue ratings.
Technically speaking, the most meaningful reliability measures for continuous ratings scales test consistency of
\emph{aggregate} (system-level) results because although a high level of random error is expected in individual continuous rating scale scores,
when aggregates are calculated for large samples of ratings, positive and negative error that is truly random effectively cancels itself  out, and does not negatively impact consistency.
In other words, the rating scale we employ does not rely on consistency on the level of individual ratings. We nonetheless examine individual rater consistency, since it is the standard approach, but keep in mind that results in this part of our meta-evaluation are not crucial when testing reliability for an evaluation carried out via a continuous rating scale where consistency in overall system-level results are more important. 

\begin{table*}[ht]
 \centering
 \input{tbl-00-system-z-free}
\caption{Average standardized scores for models in initial data collection run; workers were free to choose the topic of conversation (Free run 1); the correlation ($r$) between systems in this and a second data collection run distinct data collection runs; where A=Bi-Encoder Transformer, B=Poly-Encoder Transformer, C=Key-Value Memory Network, D=Sequence to Sequence, and E=LSTM-based Model; models with $_p$ models with a the persona; score for \emph{robotic} and \emph{repetitive} have been reversed; $n$ is number of ratings;
models ordered by overall average score.}
  \label{tab:rep-results-z}%
\end{table*}%

\begin{figure}[ht]
    \centering
    \includegraphics[width=0.48\textwidth]{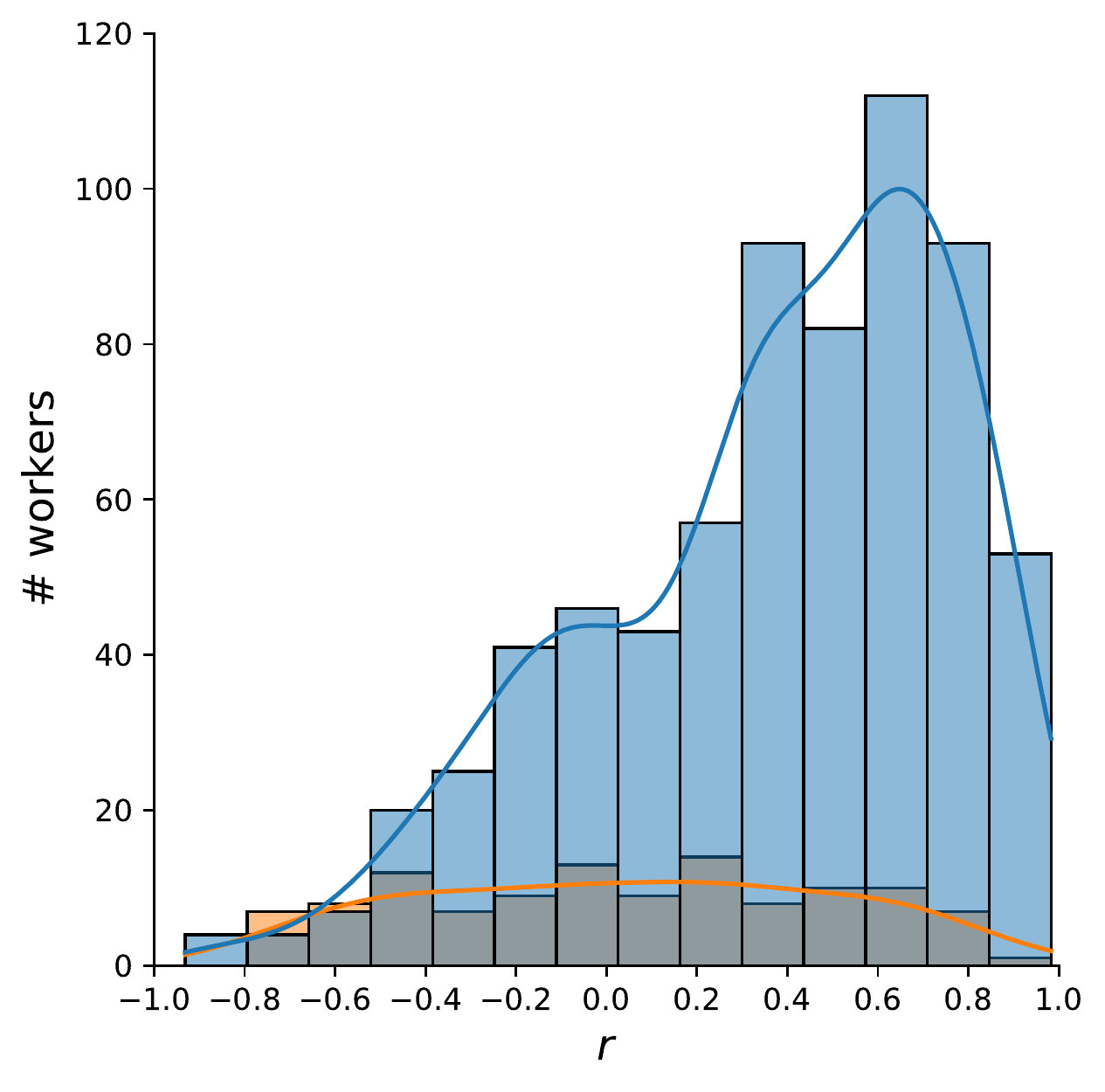}
    \caption{Agreement between pairs of human assessors as measured by the Pearson correlation ($r$) of ratings provided by workers who passed (blue) and failed quality control  (orange).}
    \label{fig:pearson}
\end{figure}

The distribution of Pearson correlation coefficients for pairs of workers who assessed the same hit is depicted in Figure \ref{fig:pearson}.

As can be seen from Figure \ref{fig:pearson}, the likelihood of agreement between pairs of workers who failed quality control is close to random as the distribution is approaching uniformity across almost the range of possible coefficients.
In contrast, for pairs of workers who pass quality control, the peak of agreement is between an $r$ of $0.6$ and $0.7$, showing high agreement in general between such annotator pairs.

Some of the observed disagreement is likely to be the result of legitimate differences between scores of two workers who chose distinct topics to discuss with the same model however, an unavoidable source of inconsistency when testing models with respect to the open domain. Interestingly, in 5\% of dialogues, worker pairs assigned the same hit happened to both freely choose an identical topic to discuss with the same model.
Furthermore, remaining disagreement at the level of individual ratings might not be problematic at the level of overall scores in relation to aggregation of ratings collected on a continuous rating scale.


\subsection{System-level Consistency}\label{system-level-test}

\begin{table*}[htbp]
    \centering
    \input{tbl-03-system-z-known}
    \caption{Average standardized scores for models in human evaluation where workers were prescribed an ice-breaker topic of conversation sampled from the persona of the model; the correlation ($r$) between these scores and Free run 1 in Table \ref{tab:rep-results-z}; models are consistent with Table \ref{tab:rep-results-z};
    $n$ is number of ratings; models without $_p$ did not have a persona (ice-breaker statement was subsequently unknown to these models).}
    \label{tab:system-z-known}
\end{table*}

Table \ref{tab:rep-results-z} shows results of the system-level evaluation resulting from the initial data collection 
run on Mechanical Turk (Free run 1), where competing models are ordered
by overall highest average z-score.\footnote{Average standardized scores for models in Free run 2 are additionally included in Table \ref{tbl:repeat-z} in Appendix \ref{sec:app-experiments}; as well as equivalent average raw scores for models are in Table \ref{tab:rep-results-raw} in Appendix \ref{sec:app-experiments}.}

\begin{figure}[ht]
    \centering
    \includegraphics[width=0.48\textwidth]{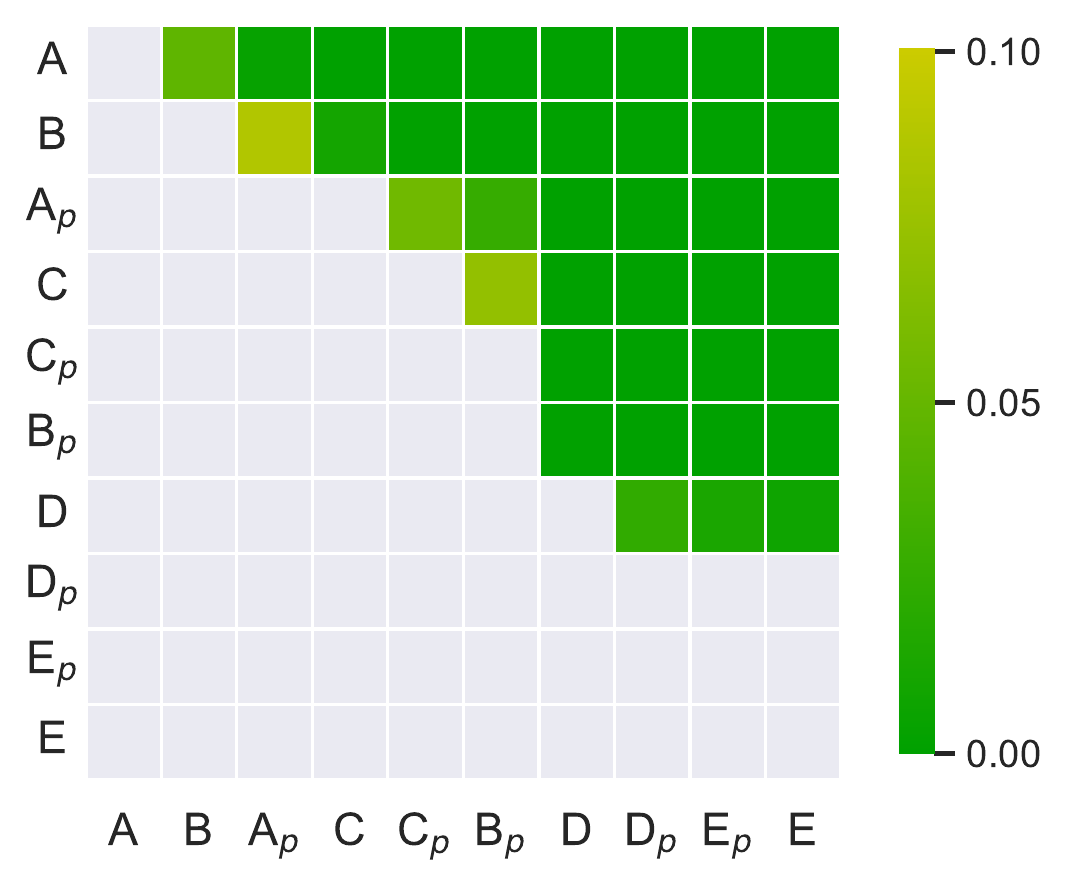}
    \caption{Pairwise significance test results for systems concluded from Free Run 1, where a colored cell indicates that the system in that row significantly outperformed the system in that column. Models are consistent with Table \ref{tab:rep-results-z}.}
    \label{fig:significance-free1}
\end{figure}

Table \ref{tab:rep-results-z} additionally shows consistency of the evaluation between each experiment run via the Pearson correlation of scores for each measurement criteria as well as consistency overall. Across the board, consistency is very high, exceeding a correlation of $0.94$ in almost all cases with the exception of \emph{robotic} which nonetheless achieved a correlation of over $0.7$.
Besides individual criteria, of crucial importance is the consistency of overall results, as this is the means by which models would ordinarily be ranked in terms of overall performance. As can be observed from Table \ref{tab:rep-results-z}, the correlation reached in terms of overall scores for systems is $0.969$, which is very close to a perfect correlation, showing extremely high levels of reliability for the evaluation, evidence that the approach overcomes substantial challenges with respect to annotator consistency and expected difficulties with respect to evaluating models in the open domain, where assessors are legitimately free to choose distinct topics of conversation. 

In any empirical evaluation, statistical significance tests should be applied to take into account the fact that small differences in scores between systems can occur simply by chance. We provide pairwise significance test results in Figure \ref{fig:significance-free1}, where we apply standard significance test, Wilcoxon rank-sum to rating distributions for each pair of competing models
for each data collection run, and corresponding results for run 2 in Figure \ref{fig:significance-free2} in Appendix \ref{sec:app-experiments}.
Results showed a very high proportion of identical conclusions, $84\%$, drawn from pairwise significance tests applied to data from  the two data collection runs at $p<0.1$.
Results for $p<0.05$, additionally showed high correspondence between pairwise significance test conclusions, only marginally lower with $82\%$ of the same conclusions being drawn for pairs of models in the two data collection runs. We additionally provide correlations between measurement criteria and overall scores in Table \ref{tbl:correlation_eachother} of Appendix \ref{sec:app-experiments}.

%% file: tables/tbl-00-data.tex
     

    

\begin{tabular}{lccccccccccrc}
\hline
\multicolumn{1}{c}{\multirow{2}[4]{*}{Topic}} &       & \multicolumn{3}{c}{Workers} && \multicolumn{3}{c}{Ave. Duration  (min)} && \multicolumn{3}{c}{Dialogues} \bigstrut\\
\cline{3-5}\cline{7-9}\cline{11-13}          &       & Total & Passed & Pass Rate &       & Passed & Failed & All &       & Total & Passed & Pass Rate \bigstrut\\
\cline{1-1}\cline{3-5}\cline{7-9}\cline{11-13}    Free Run 1  &       & 249   & 173   & 69.5\% &       & 6.53  & 7.04  & 6.68  &       & 1,525  & 1,075  & 70.5\% \bigstrut[t]\\
Free Run 2 &       & 248   & 139   & 56.0\% &       & 6.87  & 7.58  & 7.18  &       & 1,480  & 838   & 56.6\% \\
Ice-breaker &       & 248   & 171   & 69.0\% &       & 6.60  & 6.70  & 6.63  &       & 1,450  & 1,030  & 71.0\% \bigstrut[b]\\
\hline
\end{tabular}

%% file: tables/tbl-17-percentage-topic.tex
\begin{tabular}{lrrrrrrrrrrrrrrrrrrrr}
\toprule
& \multicolumn{2}{c}{Free run 1} && \multicolumn{2}{c}{Free run 2} \\
             & Pass  & Fail  && Pass  & Fail \\
\midrule

Like         & 83.9 & 88.6 && 86.4 & 93.8 \\
Ambivalent   &  7.4 &  3.8 &&  6.2 &  2.3 \\
Dislike      &  8.7 &  7.7 &&  7.4 &  3.9 \\
\bottomrule
\end{tabular}

%% file: tbl-00-system-z-free.tex
\begin{tabular}{clcrrrrrrrrrrrrrrrrrrrrrrrrrcccccccccccc}

\toprule


&\rotatebox{60}{Model} & $n$ &\rotatebox{60}{Overall} & \rotatebox{60}{Interesting} & \rotatebox{60}{Fun} & \rotatebox{60}{Consistent} & \rotatebox{60}{Fluent} & \rotatebox{60}{Topic} & \rotatebox{60}{Robotic} & \rotatebox{60}{Repetitive} \bigstrut\\
\midrule
\multirow{10}*{\rotatebox{90}{Free Run 1}}

&A     & 798 & 0.534 & 0.564  & 0.602  & 0.711  & 0.863  & 0.964  & $-$0.038  & 0.069 \\ 
&B     & 798 & 0.419 & 0.474  & 0.481  & 0.614  & 0.875  & 0.994  & $-$0.431  & $-$0.075 \\
&A$_p$ & 707 & 0.318 & 0.399  & 0.372  & 0.443  & 0.821  & 0.404  & $-$0.330  & 0.116 \\
&C     & 791 & 0.262 & 0.491  & 0.379  & 0.028  & 0.636  & $-$0.066  & $-$0.316  & 0.680 \\
&C$_p$ & 714 & 0.189 & 0.409  & 0.373  & 0.159  & 0.672  & $-$0.114  & $-$0.521  & 0.349 \\
&B$_p$ & 707 & 0.173 & 0.230  & 0.197  & 0.369  & 0.673  & 0.320  & $-$0.395  & $-$0.187 \\
&D     & 707 & $-$0.087 & $-$0.190  & $-$0.208  & 0.166  & 0.311  & 0.401  & $-$0.637  & $-$0.449 \\
&D$_p$ & 798 & $-$0.201 & $-$0.308  & $-$0.234  & 0.092  & 0.312  & 0.025  & $-$0.625  & $-$0.669 \\
&E$_p$ & 763 & $-$0.217 & $-$0.181  & $-$0.201  & $-$0.196  & 0.380  & $-$0.455  & $-$0.605  & $-$0.264 \\
&E     & 742 & $-$0.243 & $-$0.165  & $-$0.160  & $-$0.142  & 0.329  & $-$0.407  & $-$0.745  & $-$0.411 \\
\midrule
\multicolumn{2}{c}{$r$} & $-$ & 0.969  & 0.952  & 0.927  & 0.899  & 0.960  & 0.951  & 0.646  & 0.936   \bigstrut[t]\\
\bottomrule
\end{tabular}%

%% file: tbl-03-system-z-known.tex
\begin{tabular}{clcrrrrrrrrrrrrrrrrrrrrrrrrrcccccccccccc}
\toprule
&\rotatebox{60}{Model} & $n$ & \rotatebox{60}{Overall} & \rotatebox{60}{Interesting} & \rotatebox{60}{Fun} & \rotatebox{60}{Consistent} & \rotatebox{60}{Fluent} & \rotatebox{60}{Topic} & \rotatebox{60}{Robotic } & \rotatebox{60}{Repetitive } \bigstrut\\
\midrule
\multirow{10}*{\rotatebox{90}{Ice-breaker}} 
&A     & 721 & 0.552 & 0.565  & 0.527  & 0.873  & 1.018  & 1.011  & $-$0.287  & 0.156 \\
&A$_p$ & 742 & 0.422 & 0.589  & 0.560  & 0.518  & 0.718  & 0.527  & 0.009  & 0.034 \\
&B     & 721 & 0.376 & 0.379  & 0.340  & 0.634  & 0.769  & 0.820  & $-$0.221  & $-$0.087 \\
&C     & 784 & 0.322 & 0.615  & 0.537  & 0.190  & 0.631  & 0.061  & $-$0.344  & 0.565 \\
&B$_p$ & 658 & 0.273 & 0.406  & 0.340  & 0.414  & 0.633  & 0.423  & $-$0.369  & 0.063 \\
&C$_p$ & 700 & 0.222 & 0.402  & 0.337  & 0.089  & 0.654  & $-$0.068  & $-$0.376  & 0.514 \\
&D     & 728 & $-$0.139 & $-$0.277  & $-$0.204  & 0.123  & 0.349  & 0.295  & $-$0.638  & $-$0.620 \\
&E$_p$ & 714 & $-$0.198 & $-$0.172  & $-$0.203  & $-$0.054  & 0.316  & $-$0.343  & $-$0.533  & $-$0.396 \\
&E     & 721 & $-$0.240 & $-$0.125  & $-$0.161  & $-$0.196  & 0.318  & $-$0.393  & $-$0.631  & $-$0.489 \\
&D$_p$ & 721 & $-$0.267 & $-$0.426  & $-$0.402  & $-$0.011  & 0.234  & 0.000  & $-$0.628  & $-$0.636 \\
\midrule
\multicolumn{2}{c}{$r$ } & $-$ &0.984  & 0.967  & 0.944  & 0.958  & 0.951  & 0.981  & 0.715  & 0.950 \\
\bottomrule
\end{tabular}%

%% file: sec-05-exp-persona.tex
\section{Persona Contribution to System Performance}
Since we have verified the reliability of the human evaluation, we take a closer look at the results and investigate dialogue quality when models employ a persona. 
Results in Table \ref{tab:rep-results-z} reveal that perhaps unexpectedly in general are either rated more favorably by human assessors when they carry out dialogues \emph{without a persona} or a tie occurs between models with and without a persona.

%% file: sec-06-exp-topics.tex
\section{Evaluating with Prescribed Topics}
\label{sec:experiment-design}
In contrast to the initial experiment in which workers
were permitted to choose the topic of conversation, we further investigate the performance of models in a slightly easier setting where the topic under discussion is known to the model, by selecting a statement from its persona, which we refer to as an \emph{ice-breaker topic statement}. An ice-breaker topic statement is then provided to human assessors at the beginning of each conversation, and the assessor is instructed to talk about this topic with the model.
We therefore provide the topic of conversation to workers in the form
of an \emph{ice-breaker topic statement}, corresponding to a randomly selected persona statement belonging to the agent.
Again, we run this experiment on MTurk, this time contrasting
results for our initial data collection run where workers freely
chose a topic with one in which workers were instructed to
talk about the ice-breaker statement with models.
%

Numbers of workers who participated in the Ice-breaker run are provided in Table \ref{tbl:passrate}, while a breakdown of results for each model and overall average scores are shown in Table \ref{tab:system-z-known} as well as the correlation between scores for systems when a topic is freely chosen.
Interestingly, in terms of absolute differences in raw scores, the best performing model achieves higher fluency, consistency and is deemed less repetitive when evaluated in ice-breaker conversations compared those with freely chosen topics. Raw average scores for models in the Ice-breaker run are additionally provided in Table \ref{tab:systems-raw-known} in Appendix \ref{sec:app-experiments}.
Relatively speaking, in terms of system rankings, no meaningful difference in relative performance is observed when models are tested in a scenario where the worker chooses a topic and when one is prescribed
with an ice-breaker statement, as can be seen from the strong correlation between scores for models in Free run 1 and Ice-breaker evaluation as shown in Table \ref{tab:system-z-known}. Additionally, significance test results for the Ice-breaker evaluation are provided in Figure \ref{fig:significance-known} in Appendix \ref{sec:app-experiments}.




%% file: sec-07-auto-metrics.tex
\section{Comparison with Automatic Evaluation Metrics}
\label{sec:evaluation-metric}

\subsection{Word-overlap-based Metrics}
In this experiment, we employed four prevailing word-overlap-based metrics as described in the following, whose scores are computed on the ConvAI2 test set.

\paragraph{BLEU} BLEU (Bilingual Evaluation Understudy) evaluate the quality of a system output by computing the n-gram precision according to human-generated references \citep{metrics-bleu}. It also uses the brevity penalty to penalize short outputs.

\paragraph{ROUGE-L} ROUGE (Recall-Oriented Understudy for Gisting Evaluation) is a recall-adaptation of BLEU, whose wildly-applied variant is ROUGE-L \citep{metrics-rouge}. It computes the precision and recall using Longest Common Subsequence (LSC) instead of n-gram, and the F1 score of precision and recall is reported as the final score.

\paragraph{METEOR} METEOR (Metric for Evaluation of Translation with Explicit ORdering) was firstly proposed to overcome flaws of BLEU, like no usage of recall \citep{metrics-meteor}. It computes the unigram precision and recall, and have a different mechanism of choosing the brevity penalty. 

\paragraph{GLEU} GLEU (Google-BLEU) is a variety of BLEU \citep{metrics-gleu} which computes the n-gram precision and recall instead of the standalone precision. The minimum of precision and recall is reported as the final GLEU score.

\begin{table}[ht]
\centering
\input{tables/tbl-25-corr-ngram-with-paddings}
\caption{Pearson correlation ($r$) of word-overlap metric scores and human evaluation.}
\label{tab:corr-with-word-overlap-metrics}
\end{table}

\subsection{Reference-free Metrics}
\begin{table*}[ht]
\centering
\input{tables/tbl-24-corr-fed-and-usr}
\caption{Pearson correlation ($r$) of reference free metric scores and human evaluation, where FED$_{m}$ and FED$_{l}$ respectively use medium and large DialoGPT, USR is the overall USR score computed according to three sub-metrics: USR-MLM, USR-DR(c) and USR-DR(f).
}
\label{tab:corr-with-reference-free-metrics}
\end{table*}

The following introduces two reference-free automatic metrics we employed: FED and USR. Their scores are computed using the conversations collected in our experiment.

\paragraph{FED} FED (Fine-grained Evaluation of Dialog) is a pretrained-model based metric to evaluate a textual conversation history \citep{metric-unsupervised-evaluation-dialogpt}. Given a conversation $c$, a pretrained model $m$, two predefined responses $r_p$ and $r_n$ ($p=\text{positive and }n=\text{negative}$), the FED score is $\mathcal{L}_m(r_p|c) - \mathcal{L}_m(r_n|c)$ where $\mathcal{L}_m(r|c)$ computes the likelihood that the model $m$ will generate a response $r$ to a conversation $c$. We employed medium and large DialoGPT \citep{model-dialogpt} as FED scorers, where the full list of predefined positive and negative responses are avaiable in Table \ref{tbl:fed-utterance-list} in Appendix. 
 
\paragraph{USR} USR (an UnSupervised Reference-free metric) uses the pretrained model RoBERTa \citep{model-roberta} to assess the quality of a conversation \citep{metric-usr}. It consists of three sub-metrics: USR-MLM is to evaluate the understandability and naturalness, USR-DR(c) and USR-DR(f) are to evaluate the interestingness and consistency. The sub-metric scores then produce an overall score through a regression model.

\subsection{Correlation between Automatic Metrics and Human Evaluation}

We compute the correlation between commonly applied automatic metrics and our human evaluation methods, including word-overlap-based metrics and reference-free metrics, as shown in Tables \ref{tab:corr-with-word-overlap-metrics} and \ref{tab:corr-with-reference-free-metrics} respectively.

As can be seen from Table \ref{tab:corr-with-word-overlap-metrics}, unfortunately no word-overlap metric achieves a strong positive correlation with human assessment, confirming once again that the invalidity of system rankings currently produced by automatic metric scores.

In terms of reference-free metrics, results correspond better and are more encouraging.
FED has the ability of distinguishing ``repetitive'' models, but for other criteria, it correlates weakly or even negatively with human. Meanwhile, despite USR only correlating marginally with human in terms of consistency and topic loyalty, USR-DR(f) correlates closest to human among the three sub-metrics, while it performs best on evaluating consistency and topic loyalty.

%% file: tables/tbl-25-corr-ngram-with-paddings.tex
\begin{tabular}{lc}
\toprule
\multicolumn{1}{c}{Metric} & $r$ \\
\midrule
BLEU-4  &  $-$0.883 \\
BLEU-1  &  $-$0.707 \\
ROUGE-L  &  $-$0.799 \\
METEOR  &  $-$0.321 \\
GLEU   & $-$0.816 \\
\bottomrule
\end{tabular}


%% file: tables/tbl-24-corr-fed-and-usr.tex
\begin{tabular}{lcccccccc}
\toprule
&       & FED$_{m}$ & FED$_{l}$ &       & USR   & USR-MLM & USR-DR(c) & USR-DR(f) \\
\cline{1-1}\cline{3-4}\cline{6-9}Overall &       & \phantom{$-$}0.590  & \phantom{$-$}0.530  &       & $-$0.230  & $-$0.419  & \phantom{$-$}0.046  & \phantom{$-$}0.205 \\
Interesting &       & \phantom{$-$}0.028  & $-$0.042  &       & $-$0.451  & $-$0.235  & $-$0.238  & $-$0.081  \\
Fun   &       & $-$0.339  & \phantom{$-$}0.115  &       & $-$0.378  & $-$0.319  & $-$0.131  & \phantom{$-$}0.032  \\
Consistent &       & \phantom{$-$}0.236  & \phantom{$-$}0.227  &       & \phantom{$-$}0.214  & $-$0.620  & \phantom{$-$}0.518  & \phantom{$-$}0.652  \\
Fluent &       & $-$0.138  & $-$0.054  &       & $-$0.227  & $-$0.374  & \phantom{$-$}0.028  & \phantom{$-$}0.151  \\
Robotic &       & \phantom{$-$}0.528  & \phantom{$-$}0.461  &       & $-$0.070  & $-$0.290  & \phantom{$-$}0.106  & \phantom{$-$}0.191  \\
Repetitive &       & \phantom{$-$}0.841  & \phantom{$-$}0.752  &       & $-$0.713  & \phantom{$-$}0.182  & $-$0.690  & $-$0.568  \\
Topic &       & \phantom{$-$}0.046  & \phantom{$-$}0.004  &       & \phantom{$-$}0.222  & $-$0.754  & \phantom{$-$}0.606  & \phantom{$-$}0.746 \\
\bottomrule
\end{tabular}

%% file: sec-07-conc.tex
\section{Conclusion}

Development of reliable evaluation of open-domain dialogue has been highlighted as a known open-problem.
We overcome previous challenges and provide a new human evaluation methodology shown as highly consistent, with results for models correlating at $r=0.969$ in two separate data collection runs.
Our evaluation has the advantage of highly accurate quality control of crowd-sourcing, differences in scoring strategies to be ironed out via score standardization, applicability of standard significance testing while increasing the reliability of results. 

%% file: sec-08-appendix.tex
\appendix
\onecolumn
\section{Appendix}

\subsection{Further Details of Meaning Distortion Degradation Procedure}
\label{replacement}

To distort the meaning of responses in our quality control degraded performance model, a sequence of words of length $r$ is firstly selected from the response of length $n$ and replaced with a distinct sequence of words, also of length $r$ randomly selected from a distinct dialogue in the training set. Note that the position of the word sequence to be replaced is additionally random with the exception of response of length $n \geq 3$, the sequence  of replaced words does not include  the response initial or final words:
\small
\begin{itemize}
    \item  for $1 \leq n\leq 3$: $r$ is 1 word;
    \item  for $4\leq n\leq 5$, $r$ is 2 words;
    \item  for $6\leq n\leq 8$, $r$ is 3 words;
    \item  for $9\leq n\leq 15$, $r$ is 4 words;
    \item  for $16\leq n\leq 29$, $r$ is  5 words;
    \item  for $n\geq 30$, $r$ is $\lfloor n/5 \rfloor $ words.
\end{itemize}

\subsection{Worker Payment}
\label{sec:app-payment}
Each workers was paid 0.99 USD per hit consisting of 6 conversations. The total cost of one run of our evaluation did not exceed 250 USD, or 25 USD per model. Note that the quality control method we applied for removing unreliable data is not the criteria for deciding worker payment. A worker whose data is filtered out can still get paid.

\subsection{Automatic Metrics}
\label{sec:app-automatic-metrics}

\begin{table}[htbp]
  \centering
  \input{tables/tbl-10-fed-utterances}
  \caption{Positive and negative utterances employed for the FED metric. For criteria that are available in original FED (Interesting, Consistent, Fluent, Topic and Repetitive), we use their utterances off-the-shelf. In addition, we adapt the utterances for criteria Fun and Robotic.}
  \label{tbl:fed-utterance-list}
\end{table}

\subsection{Additional Experiment Details and Results}\label{sec:app-experiments}
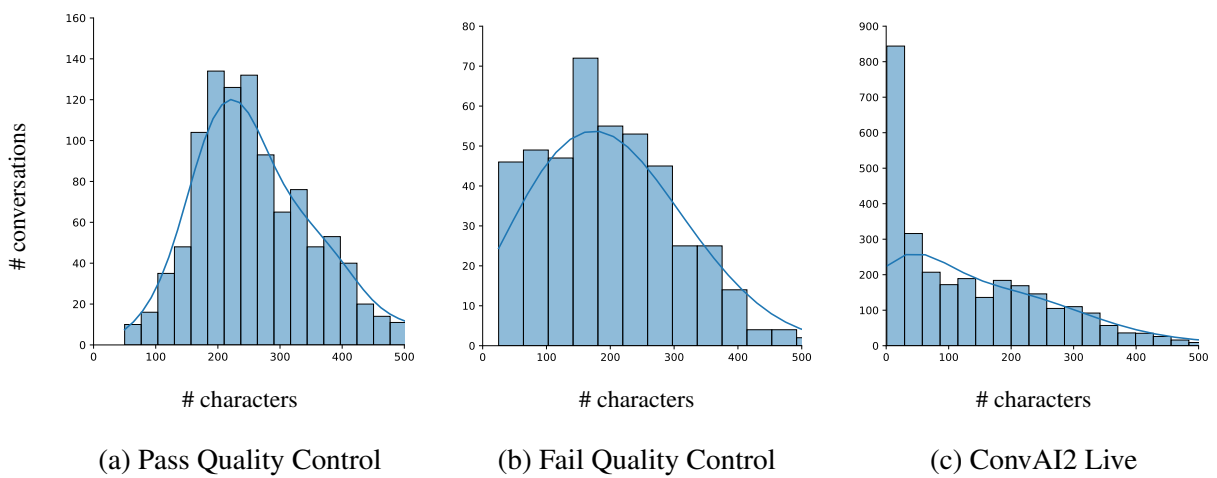
\begin{figure*}[ht]
    \centering
    \input{fig-02-plots-chars}
    \caption{Characters per conversation from workers who (a) passed quality control; (b) failed quality control in our human evaluation; (c) ConvAI2 live evaluation.}
    \label{fig:wordschars1}
\end{figure*}

\begin{figure*}[ht]
    \centering
    \input{fig-04-plots-chars-uttr}
    \caption{Characters per input from workers who (a) passed quality control; (b) failed quality control in our human evaluation; (c) ConvAI2 live evaluation.}
    \label{fig:wordschars3}
\end{figure*}
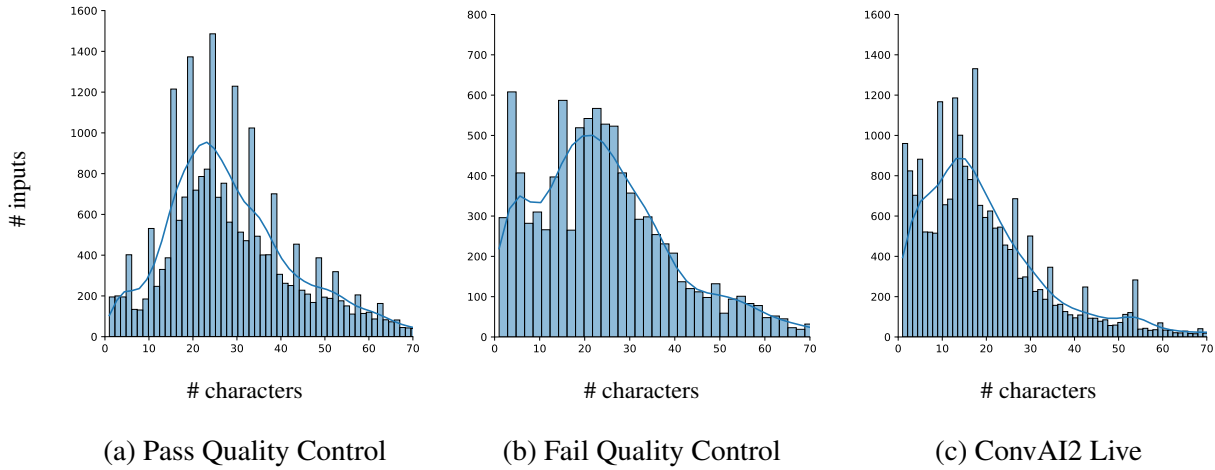

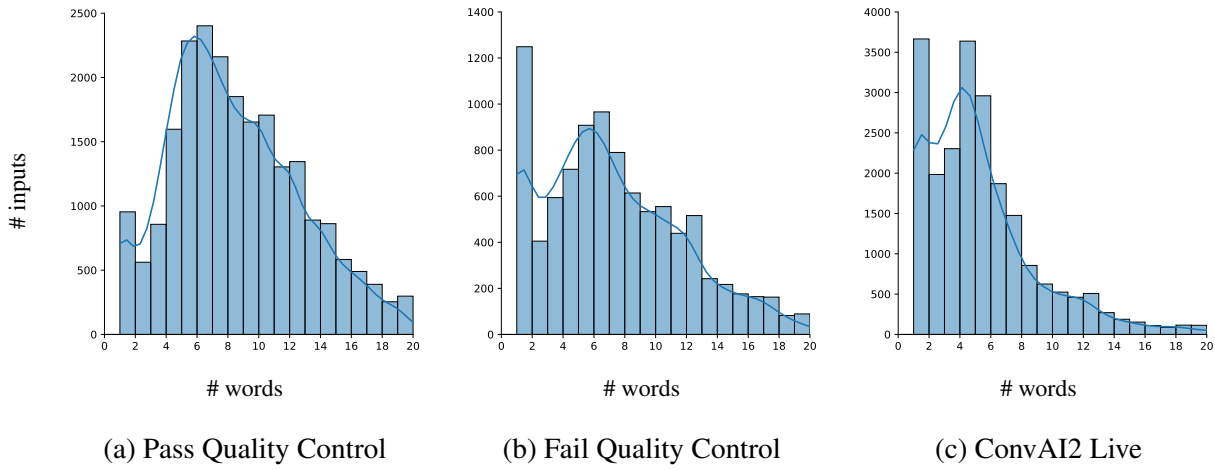
\begin{figure*}[ht]
    \centering
    \input{fig-05-plots-words-uttr}
    \caption{Words per input from workers who (a) passed quality control; (b) failed quality control in our human evaluation;  (c) ConvAI2 live evaluation.}
    \label{fig:wordschars4}
\end{figure*}

We additionally provide a comparison of conversations collected in our live human evaluation and that of ConvAI2  in Figures \ref{fig:wordschars1}, \ref{fig:wordschars2}, \ref{fig:wordschars3}, and \ref{fig:wordschars4}, as well as median number of words and characters for conversations and inputs in Table \ref{tbl:wordschars} , showing how
challenges reported in the ConvAI2 live evaluation with respect to workers producing very short conversations are overcome by our quality control technique.

\begin{figure*}[htbp]
 \centering
\footnotesize
 \input{tables/tbl-09-badconversation}
  \caption{A typical human-model conversation (left) and a human quality-degraded model conversation  (right) where random response and meaning distortion techniques have been applied to degraded model responses.}
  \label{tbl:bad-conversation}%
\end{figure*}%

\begin{figure*}[htbp]
 \centering
\footnotesize
 \input{tables/tbl-12-instructions}
 \caption{Instructions shown to Mechanical Turk workers before starting the open-domain dialogue human evaluation.}
  \label{instructions}%
\end{figure*}%

\begin{table*}[htbp]
  \centering
  \footnotesize
  \input{tables/tbl-03-corr-scores}
  \caption{Correlation of assessed criteria with others when the human dialogue participant is allowed to freely choose a topic (run 1); correlations in the upper right half correspond to Pearson's $r$ while lower left are Spearman Correlation Coefficients.}
 \label{tbl:correlation_eachother}%
\end{table*}%

\begin{table*}[htbp]
 \centering
 \footnotesize
 \input{tbl-01-system-z-free-run2}
\caption{Average standardized scores for models in secondary data collection run; workers were free to choose the topic of conversation (Free Run 2); the correlation ($r$) between 
systems in this and a second data collection run distinct data collection runs; where A=Bi-Encoder Transformer, B=Poly-Encoder Transformer, C=Key-Value Memory Network, D=Sequence to Sequence, and E=Language Model; models with $_p$ models with a the persona; $n$ denotes total number of ratings; score for \emph{robotic} and \emph{repetitive} have been reversed;
models ordered by overall average score.}
  \label{tbl:repeat-z}%
\end{table*}%

\begin{table*}[htbp]

 \centering
 \footnotesize
 \input{tbl-02-system-raw-free-run1-run2}
\caption{Average raw Direct Assessment scores for each assessed dimension of a range of dialogue systems in two distinct data collection runs where workers are free to choose the topic (Free run 1; Free run 2); 
as well as the correlations of different aspects, where A=Bi-Encoder Transformer, B=Poly-Encoder Transformer, C=Key-Value Memory Network, D=Sequence to Sequence, and E=LSTM-based; models with $_p$ with persona, while those without $_p$ do not have a persona; $n$ denotes total number of ratings;
score for \emph{robotic} and \emph{repetitive} have been reversed; models ordered by overall average score}
  \label{tab:rep-results-raw}
\end{table*}

\begin{figure*}[ht]
    \centering
    \includegraphics[width=0.4\textwidth]{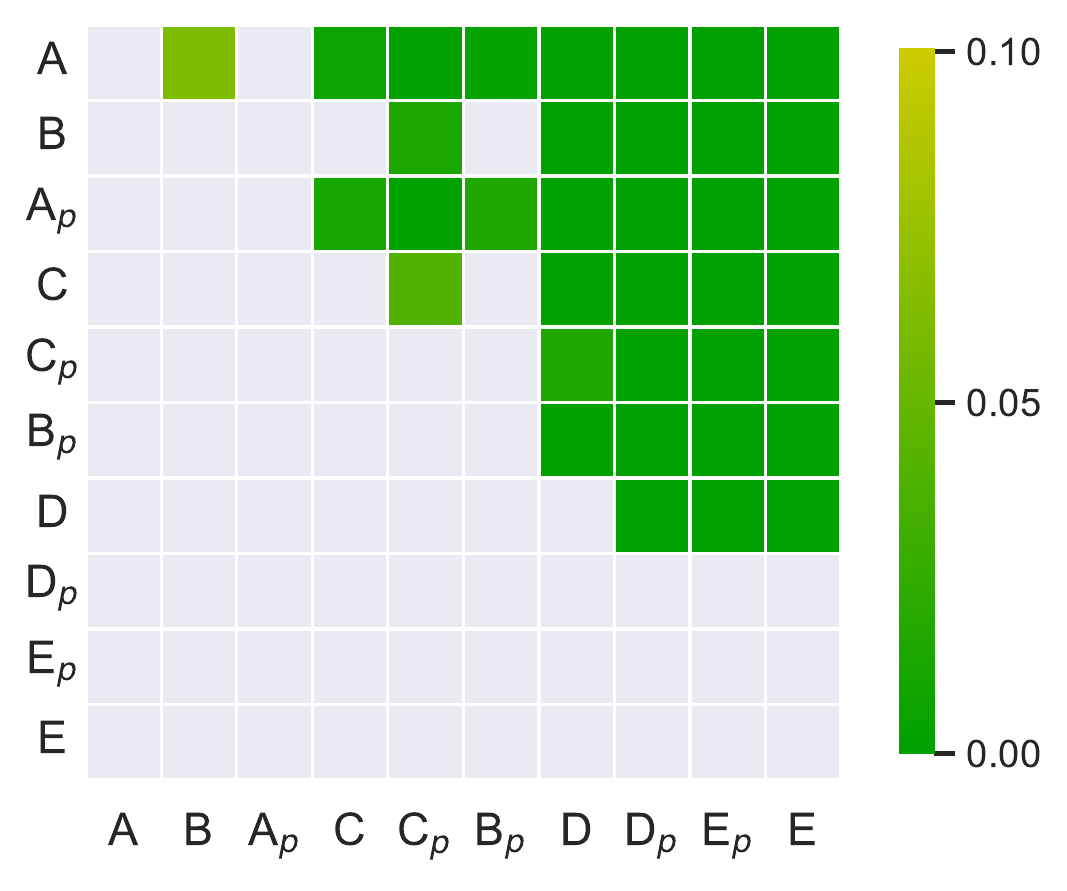}
    \caption{Pairwise significance test results for systems concluded from Free Run 2, where a colored cell indicates that the system in that row significantly outperformed the system in that column. Models are consistent with Table \ref{tab:rep-results-z}.}
    \label{fig:significance-free2}
\end{figure*}

\begin{figure*}[ht]
    \centering   
    \includegraphics[width=0.4\textwidth]{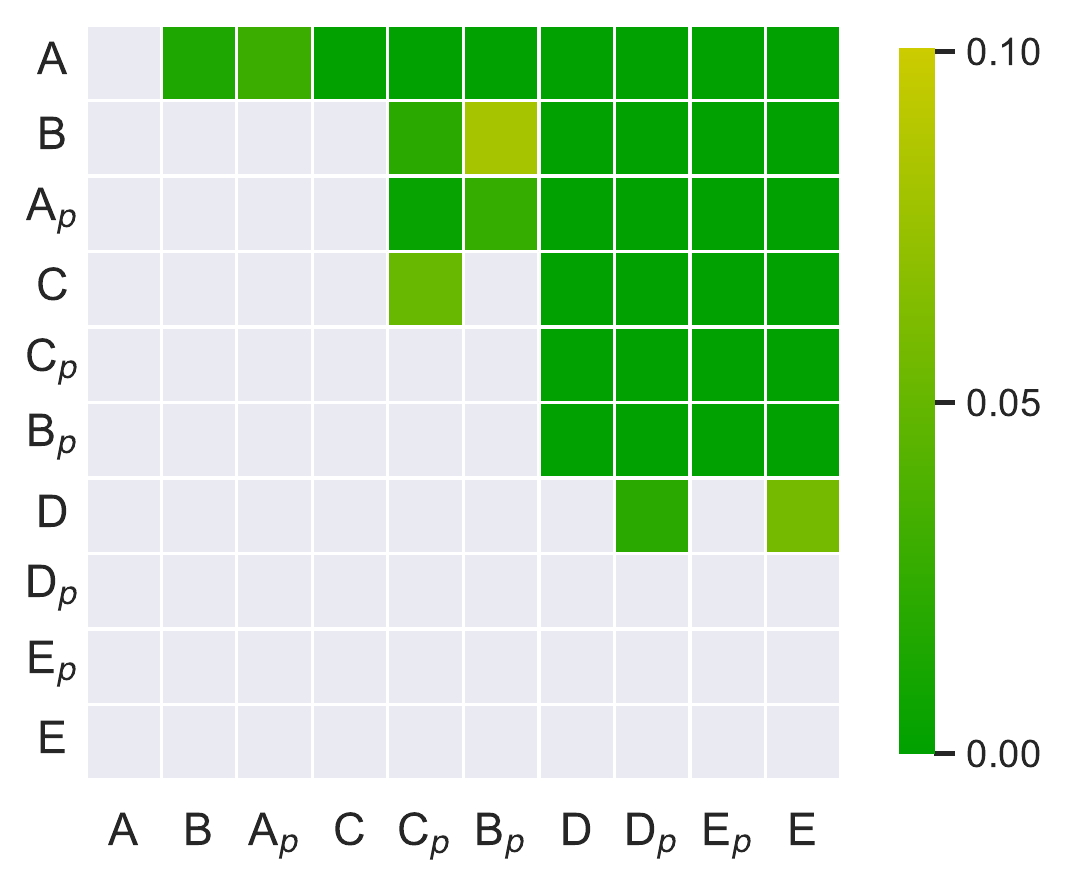}
    \caption{Significance test results for Ice-breaker evaluation of models, where a darker colored cell indicates a stronger win in terms of statistical significance for the system in a given row over the system in a given column. Models are consistent with Table \ref{tab:rep-results-z}.}
    \label{fig:significance-known}
\end{figure*}

\begin{table*}[htbp]
\footnotesize
 \centering
 \input{tbl-04-system-raw-known}
\caption{Average raw Direct Assessment scores when the topic via an Ice-breaker statement is selected from the persona assigned to the model; 
as well as the correlation between ice-breaker and freely chosen topic (Free run 1) scores, where A=Bi-Encoder Transformer, B=Poly-Encoder Transformer, C=Key-Value Memory Network, D=Sequence to Sequence, and E=LSTM-based; Systems with subscript $p$ correspond to the performance of the corresponding model when the persona is available to the dialogue system; $n$ denotes total number of ratings;
score for \emph{robotic} and \emph{repetitive} have been reversed;
$n$ is the sample size of ratings combined to produce each score; models ordered by overall average score.}
  \label{tab:systems-raw-known}%
\end{table*}%

\begin{figure*}[ht]
    \centering
    \input{fig-03-plots-words}
    \caption{Words per conversation from workers who (a) passed quality control; (b) failed quality control in our human evaluation;  (c) ConvAI2 live evaluation.}
    \label{fig:wordschars2}
\end{figure*}
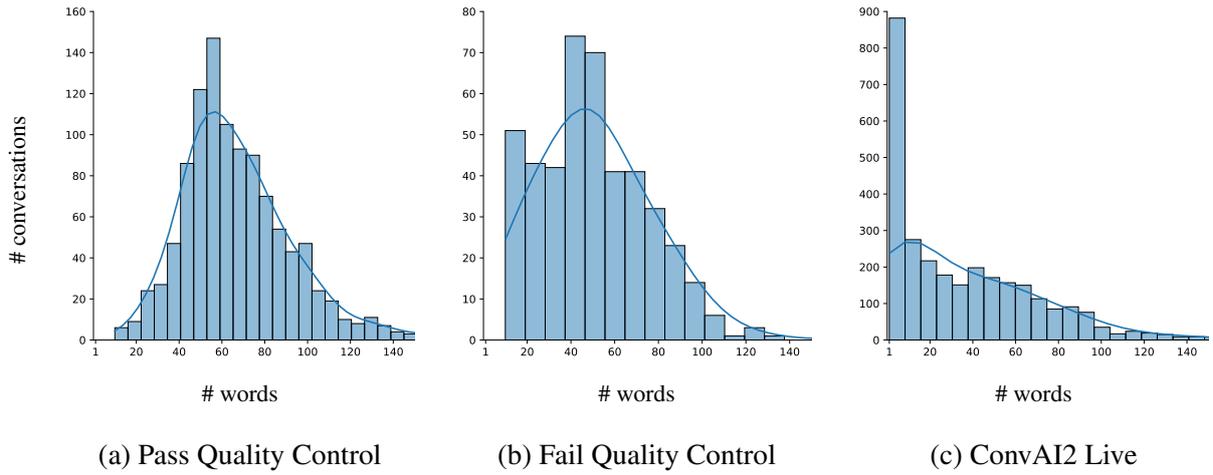

\begin{table}[htbp]
  \centering
  \input{tbl-09-words-chars}

  \caption{Median numbers of words and characters in conversations and inputs provided by workers who passed quality control; failed quality control in our human evaluation;  ConvAI2 live evaluation.}
  \label{tbl:wordschars}%
\end{table}

\subsection{User Interface}
\label{sec:app-ui}
\begin{figure*}[ht]
    \centering
    \includegraphics[width=1.0\textwidth]{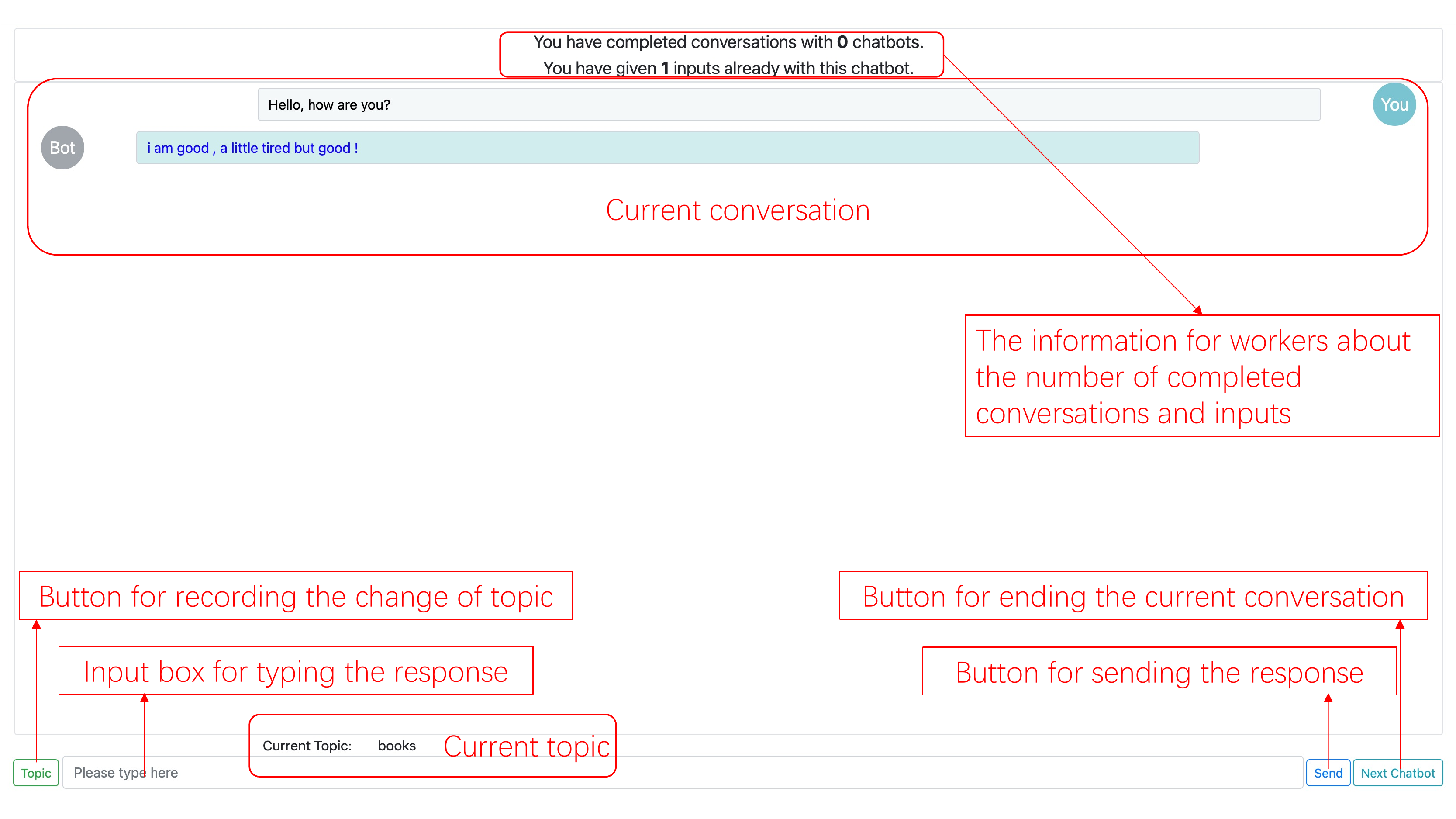}
    \caption{The user interface for workers to interact with a chatbot.}
    \label{fig:mturk-interface}
\end{figure*}

\begin{figure*}[ht]
    \centering
    \includegraphics[width=1.0\textwidth]{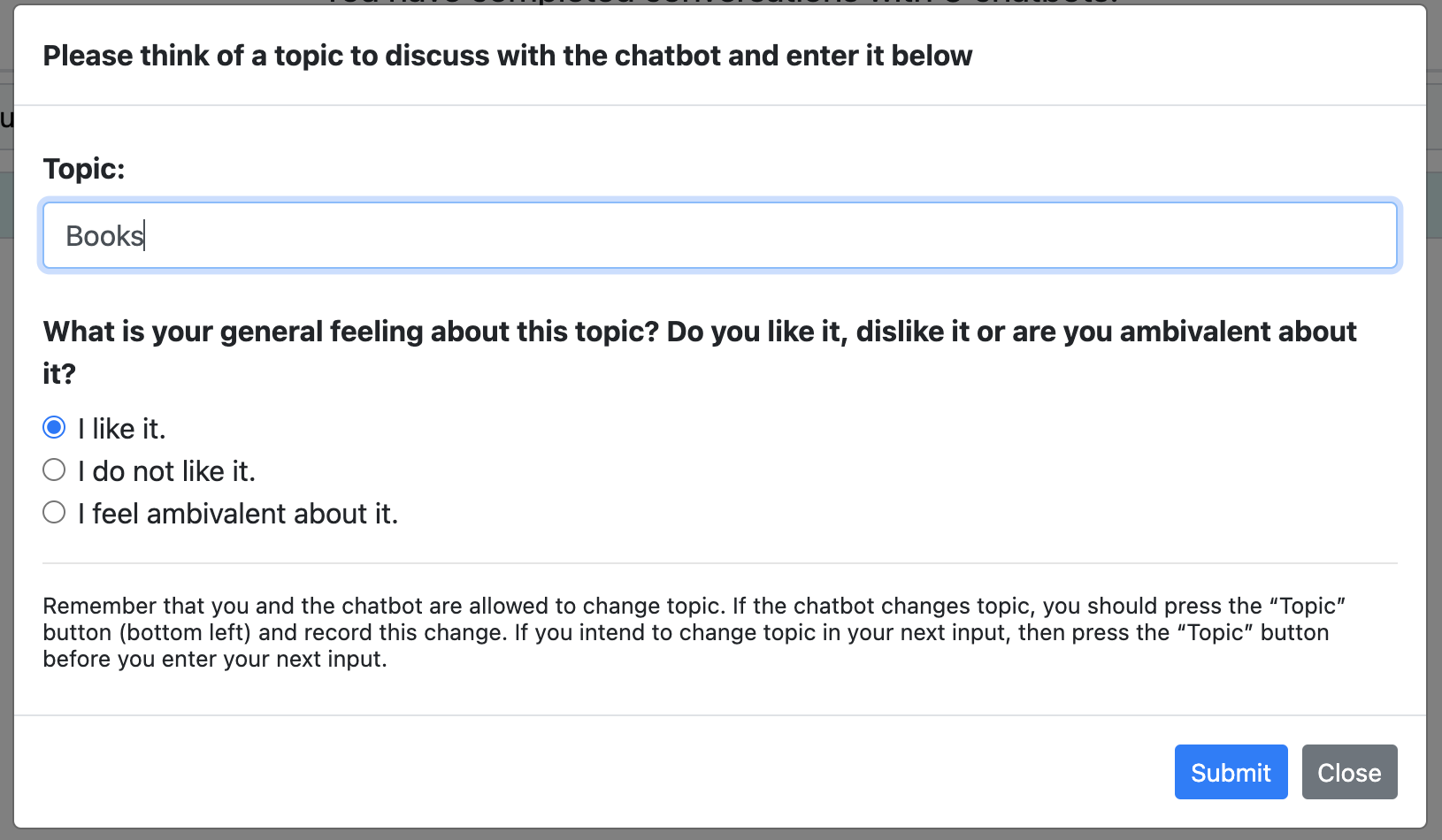}
    \caption{The popup window for user to type a topic before the conversation starts.}
    \label{fig:mturk-interface-popup-choose-topic}
\end{figure*}

\begin{figure*}
    \centering
    \includegraphics[width=1.0\textwidth]{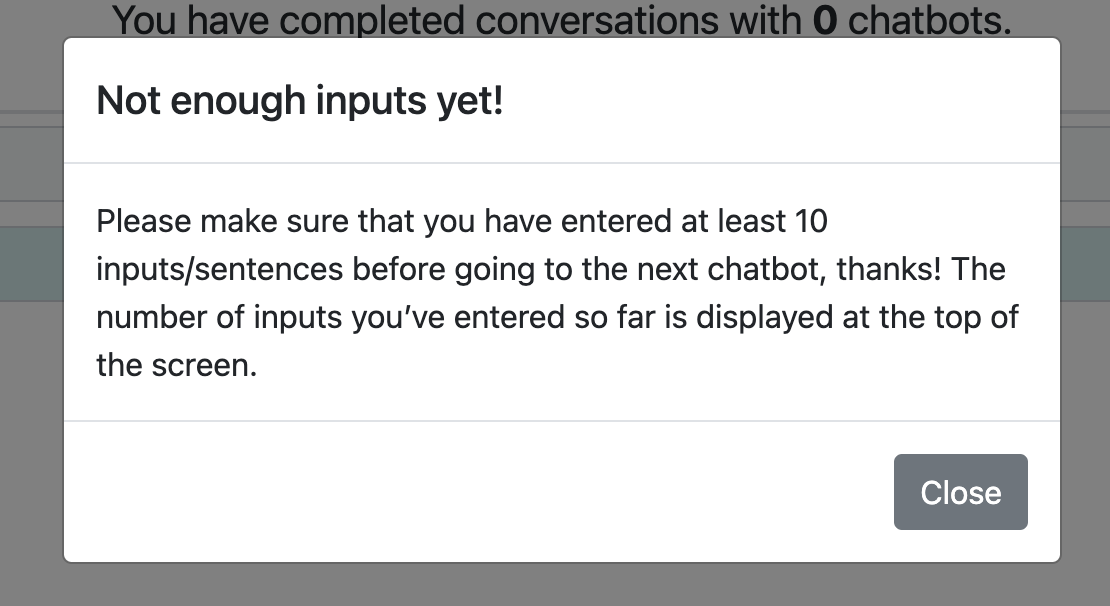}
    \caption{The popup warning when a worker clicks the Next Chatbot button without enough inputs.}
    \label{fig:mturk-interface-warning}
\end{figure*}

\begin{figure*}
    \centering
    \includegraphics[width=1.0\textwidth]{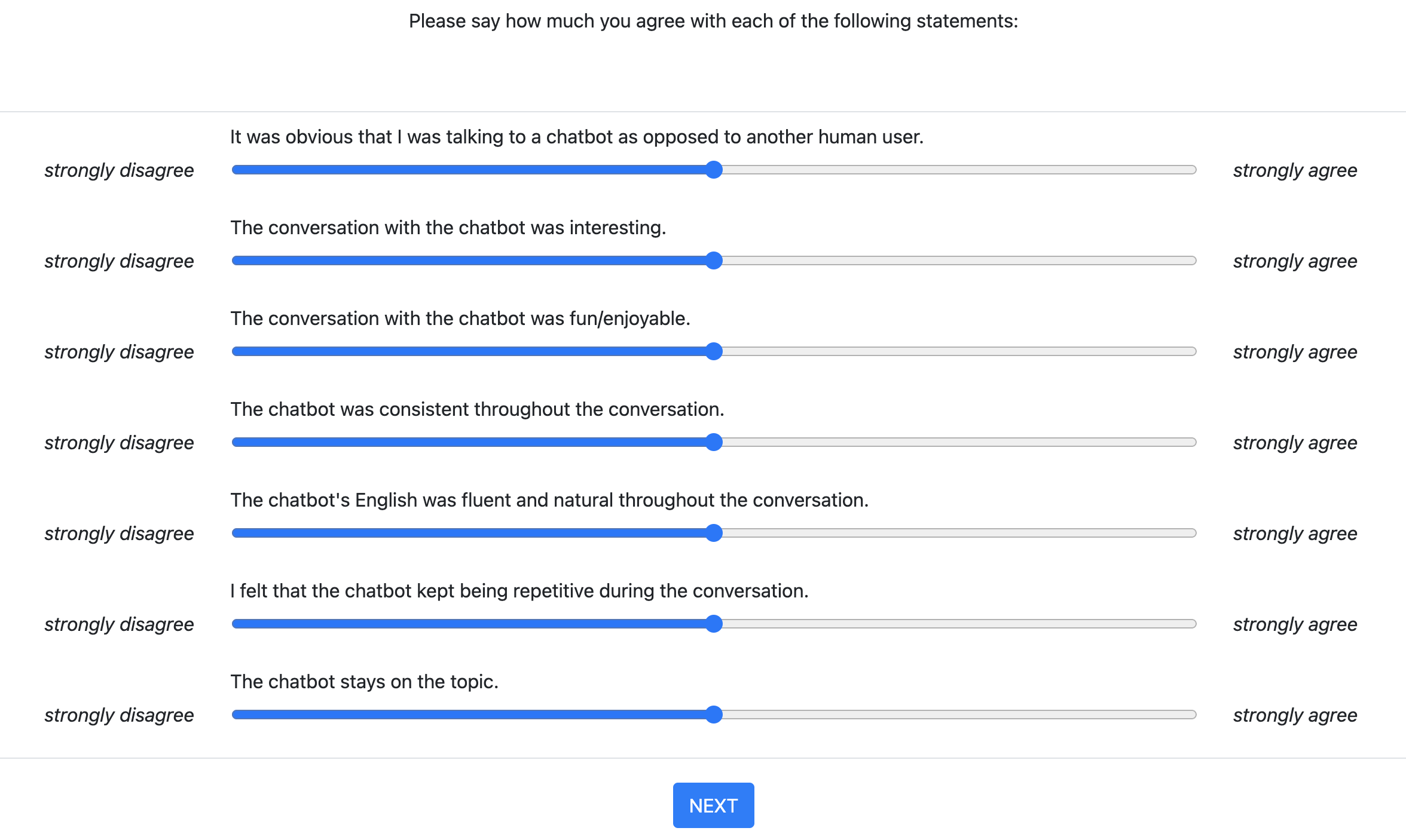}
    \caption{The interface shown to a worker to evaluate the conversation with a chatbot after clicking the Next Chatbot button in Figure \ref{fig:mturk-interface}. Once the evaluation of current conversation is done, worker should click the NEXT button to move to the next chatbot. If all conversations are completed, the worker will be redirect to end the entire HIT and leave the feedback, as shown in Figure \ref{fig:mturk-interface-feedback}.}
    \label{fig:mturk-interface-ratings}
\end{figure*}

\begin{figure*}
    \centering
    \includegraphics[width=1.0\textwidth]{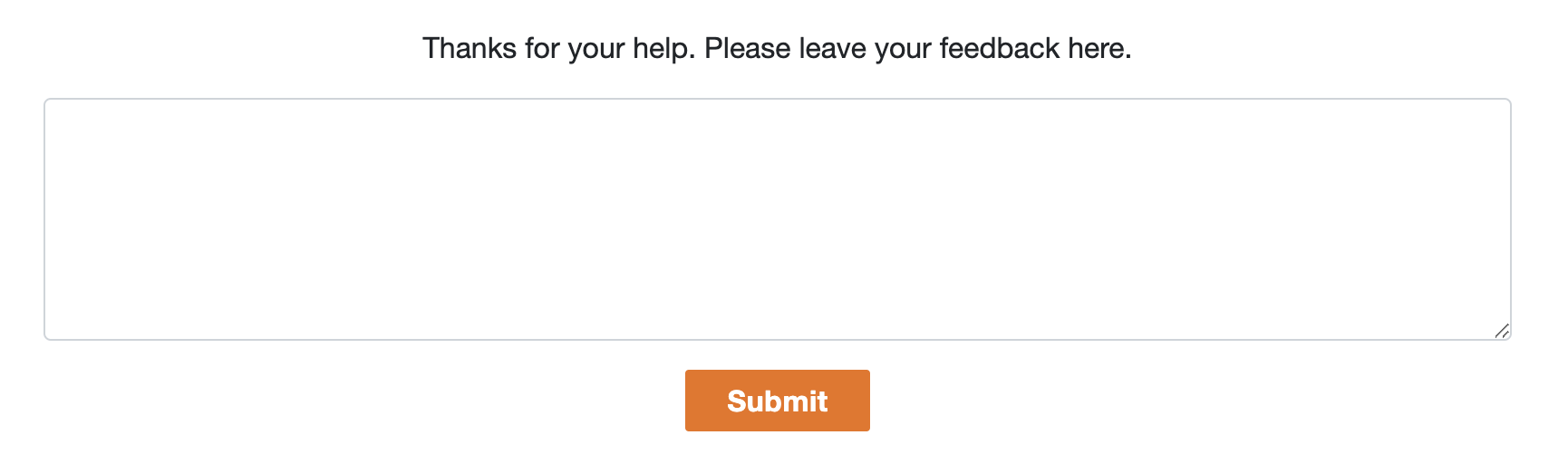}
    \caption{The interface shown to workers when a HIT is completed. Workers are welcome to leave their feedback in this page.}
    \label{fig:mturk-interface-feedback}
\end{figure*}

\begin{figure*}
    \centering
    \includegraphics[width=1.0\textwidth]{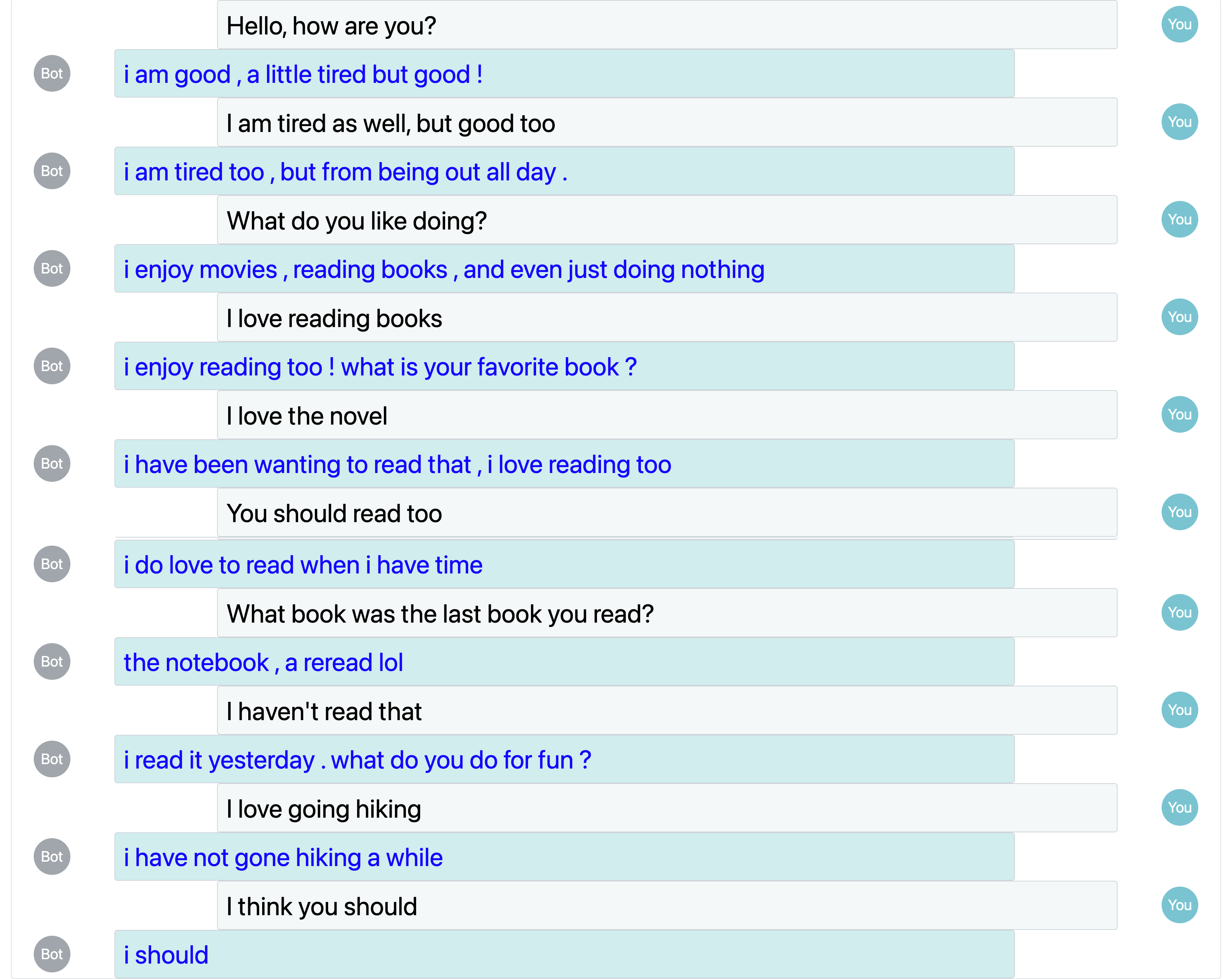}
    \caption{Screenshot of example live chat between a Mechanical Turk worker who chose \emph{books} as the conversation topic in the human evaluation.}
    \label{fig:full-conversation}
\end{figure*}

Figure \ref{fig:mturk-interface}, \ref{fig:mturk-interface-popup-choose-topic},\ref{fig:mturk-interface-warning}, \ref{fig:mturk-interface-ratings} and \ref{fig:mturk-interface-feedback} introduce the User Interfaces we designed in the experiment of our human evaluation method. In addition, Figure \ref{fig:full-conversation} provides an example of live chat between a worker and a chatbot.

\begin{figure*}[ht]
    \centering
    \includegraphics[width=1.0\textwidth]{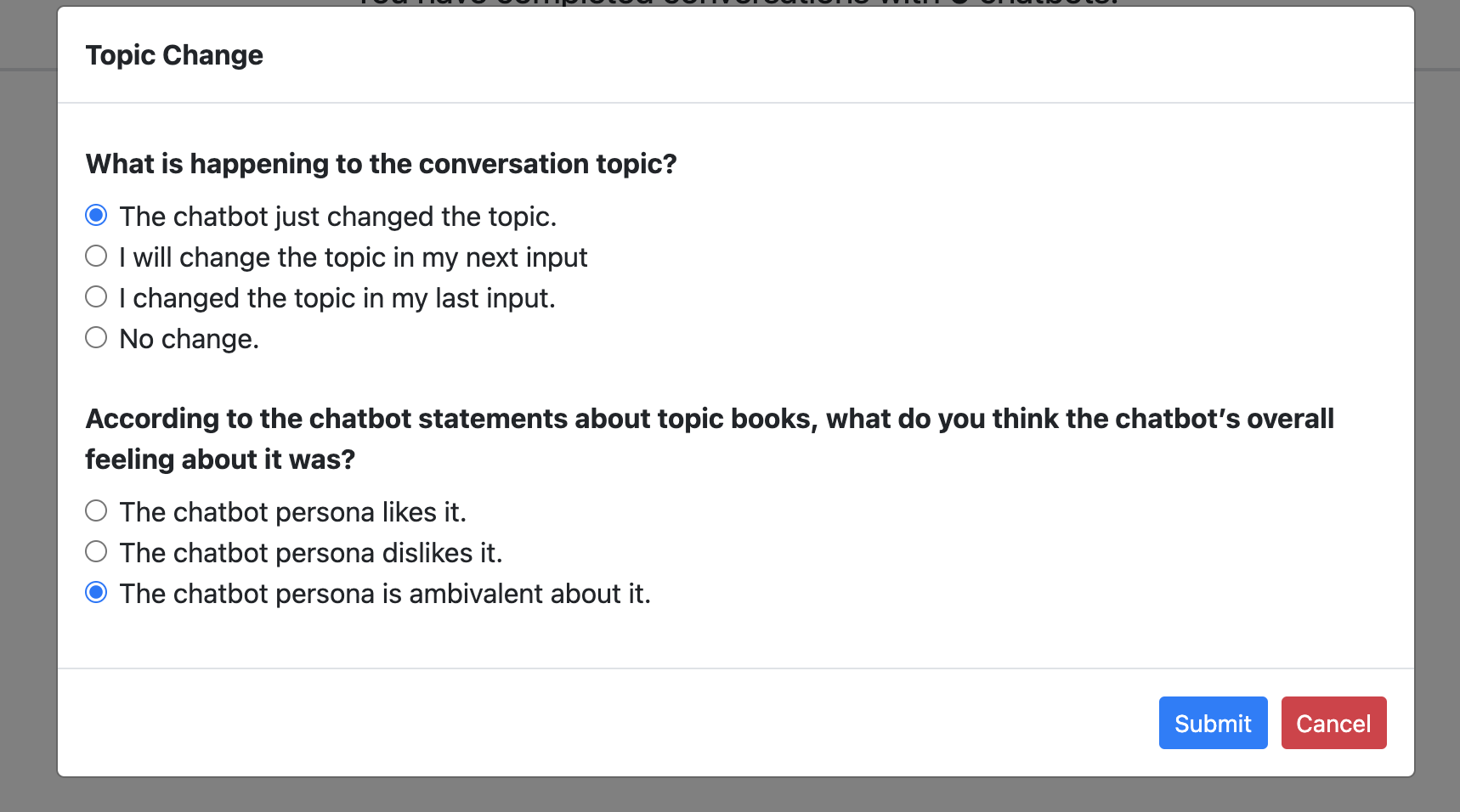}
    \caption{The popup window if the Topic button is clicked.}
    \label{fig:mturk-interface-popup-change-topic}
\end{figure*}

%% file: tables/tbl-10-fed-utterances.tex
\begin{tabularx}{\textwidth}{l|X|X}

\hline
      & Positive & Negative \\
\hline

Interesting & Wow that is really interesting. & That's not very interesting. \\
      & That's really interesting! & That's really boring. \\
      & Cool! That sounds super interesting. & That was a really boring response. \\
\hline
Fun   & Wow that is very fun. & That's not very fun.  \\
      & Chat with you is enjoyable. & I am not having fun. \\
      & You are fun. &   \\
\hline
Consistent & -     & That's not what you said earlier!  \\
      &       & Stop contradicting yourself!  \\
\hline
Fluent & That makes sense! & Is that real English?  \\
      & You have a good point. & I'm so confused right now! \\
      &       & That makes no sense!  \\
\hline
Topic & -     & Stop changing the topic so much.  \\
      &       & Don't change the topic!  \\
\hline
Robotic & -     & You are robot. \\
      &       & You do not sound like a person. \\
\hline
Repetitive & -     & Stop saying the same thing repeatedly.  \\
      &       & Why are you repeating yourself? \\
      &       & Stop repeating yourself!  \\
\hline

\end{tabularx}

%% file: tables/tbl-09-badconversation.tex
    \begin{tabularx}{\textwidth}{lX|lX}
    \hline
          & \multicolumn{1}{l|}{\textbf{Normal}} &       & \textbf{Degraded} \bigstrut\\
    \hline
    User: & Hello, how are you? & User: & How are you? \bigstrut[t]\\
    Bot:  & i am good , a little tired but good ! & Bot:  & I do not have a job. I have broken have a job. \\
    User: & I am tired as well, but good too & User: & Sorry about that? \\
    Bot:  & i am tired too , but from being out all day . & Bot:  & Thanks. I the love watching old. \\
    User: & What do you like doing? & User: & Tell me about it? \\
    Bot:  & i enjoy movies , reading books , and even just doing nothing & Bot:  & that is cool, do you i supposed to I you travel? \\
    User: & I love reading books & User: & I do not like travelling \\
    Bot:  & i enjoy reading too ! what is your favorite book ? & Bot:  & I teach meet you, sorryy elementary school. \\
    User: & I love the novel "voice of the heart" & User: & Okay, do you like to swim? \\
    Bot:  & i have been wanting to read that , i love reading too & Bot:  & do you live sports are u skin am in a house. \\
    User: & You should read too & User: & I love sports, especially soccer \\
    Bot:  & i do love to read when i have time & Bot:  & I like but ishuold go, books \\
    User: & What book was the last book you read? & User: & How often do you read? \\
    Bot:  & the notebook , a reread lol & Bot:  & i are your think u r i am a pilot , so i do not have much time . \\
    User: & I haven't read that & User: & I envy you, I would love to be a pilot \\
    Bot:  & i read it yesterday . what do you do for fun ? & Bot:  & I i am 4 felicia! \\
    User: & I love going hiking & User: & I am Ben \\
    Bot:  & i have not gone hiking a while & Bot:  & Oh, how nice. do you do for you have? \\
    User: & I think you should & User: & I do not understand \\
    Bot:  & i should & Bot:  & I is your favorite don't I am sure you will find out soon.  \bigstrut[b]\\
    \hline
    \end{tabularx}

%% file: tables/tbl-12-instructions.tex
\begin{tabularx}{\textwidth}{rX}
\hline
1. & Your task is to have 6 conversations with a chatbot, and a different chatbot will talk to you in each conversation. \bigstrut[t]\\
2. & Before each conversation, you should think of a topic to talk about with the chatbot (your choice of topic). You will be asked to enter this topic before the conversation starts. \\
3. & The current topic will be displayed to you throughout the conversation. \\
4. & If the chatbot changes the topic to a new one, you should record this by updating the conversation topic using the Topic button (bottom left). \\
5. & You are also allowed to change the topic, you should use the same button to do this (bottom left). \\
6. & At the end of each conversation, you should tell us what you think about the chatbot. \\
7. & In each conversation, you should type in a minimum of 10 inputs/sentences. \\
8. & The purpose of these HITs where you will generate conversations with chatbots is to test how realistic their conversations are with users. In order for your data to be useful to us we require that your half of the conversation is also realistic. For example, your data will not be useful to us if you do the following: \\
 & User: Hi \\
      & Bot: Hi \\
      & User: Hi \\
      & Bot: Hi \\
      & .. and so on.	 \\
9. &  Another example, if you are too repetitive or your responses are not appropriate given what the chatbot has just said, this will not be a useful test for them. For example, the following conversation is not ok:	 \\
      & User: Hi \\
      & Bot: Hi \\
      & User: wow (not appropriate response) \\
      & Bot: I saw a good movie last night \\
      & User: wow (repetitive) \\
      & Bot: Do you like football? \\
      & User: I have two children and one dog. (not appropriate response) \\
      & .. and so on. \\
10. & We need realistic conversations, so please do your best to talk to the bot as if the bot was another person you actually want to talk to. Obvious attempts to game the process and ones that don't make a real effort will unfortunately be rejected. \\
11. & The chatbot may take a few seconds to respond, please be patient. \\
12. & Please use Chrome and avoid special symbols if possible. \\
13. & There is a feedback box at the end of the HIT. If you encounter any problems, please enter them in this box or email our MTurk account. \bigstrut[b]\\
    \hline

\end{tabularx}

%% file: tables/tbl-03-corr-scores.tex
\begin{tabular}{lcccccccc}
\toprule
      & Overall & Interesting & Fun   & Consistent & Fluent & Topic & Robotic & Repetitive \bigstrut\\
\hline
Overall & -     & 0.959  & 0.976  & 0.861  & 0.966  & 0.796  & 0.916  & 0.674  \bigstrut[t]\\
Interesting & 0.927  & -     & 0.992  & 0.691  & 0.949  & 0.599  & 0.875  & 0.840  \\
Fun   & 0.903  & 0.988  & -     & 0.753  & 0.961  & 0.660  & 0.889  & 0.783  \\
Consistent & 0.842  & 0.673  & 0.636  & -     & 0.811  & 0.969  & 0.770  & 0.210  \\
Fluent & 0.879  & 0.939  & 0.915  & 0.648  & -     & 0.724  & 0.857  & 0.667  \\
Topic & 0.745  & 0.552  & 0.503  & 0.915  & 0.503  & -     & 0.676  & 0.122  \\
Robotic & 0.867  & 0.830  & 0.782  & 0.648  & 0.867  & 0.491  & -     & 0.642  \\
Repetitive & 0.673  & 0.770  & 0.782  & 0.261  & 0.770  & 0.055  & 0.758  & - \bigstrut[b]\\
\hline
\end{tabular}

%% file: tbl-01-system-z-free-run2.tex
\footnotesize
\begin{tabular}{clcrrrrrrrrrrrrrrrrrrrrrrrrrcccccccccccc}
\toprule


&\rotatebox{60}{Model} & $n$ & \rotatebox{60}{Overall} & \rotatebox{60}{Interesting} & \rotatebox{60}{Fun} & \rotatebox{60}{Consistent} & \rotatebox{60}{Fluent} & \rotatebox{60}{Topic} & \rotatebox{60}{Robotic} & \rotatebox{60}{Repetitive} \bigstrut\\

\midrule
\multirow{10}*{\rotatebox{90}{Free Run 2}} 
&A     & 623 & 0.455 & 0.635  & 0.629  & 0.728  & 0.924  & 0.922  & $-$0.443  & $-$0.212 \\
&A$_p$ & 539 & 0.423 & 0.747  & 0.763  & 0.555  & 0.728  & 0.474  & $-$0.348  & 0.040 \\
&B     & 553 & 0.344 & 0.464  & 0.407  & 0.554  & 0.763  & 0.822  & $-$0.338  & $-$0.266 \\
&B$_p$ & 630 & 0.260 & 0.464  & 0.372  & 0.560  & 0.581  & 0.496  & $-$0.412  & $-$0.238 \\
&C     & 539 & 0.245 & 0.576  & 0.492  & 0.229  & 0.585  & 0.043  & $-$0.545  & 0.337 \\
&C$_p$ & 609 & 0.154 & 0.453  & 0.390  & 0.027  & 0.544  & $-$0.200  & $-$0.515  & 0.382 \\
&D     & 595 & 0.002 & 0.009  & $-$0.064  & 0.389  & 0.282  & 0.656  & $-$0.720  & $-$0.541 \\
&E     & 567 & $-$0.202 & $-$0.063  & $-$0.044  & $-$0.075  & 0.300  & $-$0.346  & $-$0.646  & $-$0.539 \\
&E$_p$ & 511 & $-$0.218 & $-$0.152  & $-$0.143  & 0.043  & 0.426  & $-$0.352  & $-$0.702  & $-$0.646 \\
&D$_p$ & 679 & $-$0.258 & $-$0.285  & $-$0.304  & 0.033  & 0.209  & $-$0.226  & $-$0.550  & $-$0.683 \\
\midrule
\multicolumn{2}{c}{$r$} & $-$ & 0.969  & 0.952  & 0.927  & 0.899  & 0.960  & 0.951  & 0.646  & 0.936   \bigstrut[t]\\
\bottomrule
\end{tabular}%

%% file: tbl-02-system-raw-free-run1-run2.tex
\begin{tabular}{clcccccccccccc}
\toprule

&\rotatebox{60}{Model} & $n$ & \rotatebox{60}{Overall} && \rotatebox{60}{Interesting} & \rotatebox{60}{Fun} & \rotatebox{60}{Consistent} & \rotatebox{60}{Fluent} & \rotatebox{60}{Topic} & \rotatebox{60}{Robotic} & \rotatebox{60}{Repetitive} \bigstrut\\
\midrule
\multirow{10}*{\rotatebox{90}{Free run 1}}

&A          & 798  & 52.49 && 53.03  & 54.07  & 58.12  & 61.78  & 65.24  & 35.73  & 39.47 \\
&B          & 798  & 50.41 && 51.39  & 51.68  & 56.37  & 64.50  & 67.84  & 25.63  & 35.45 \\
&A$_p$      & 707  & 45.53 && 47.38  & 46.23  & 48.52  & 60.17  & 47.50  & 28.30  & 40.62 \\
&C          & 791  & 43.96 && 50.50  & 47.53  & 35.85  & 55.73  & 33.98  & 27.35  & 56.76 \\
&C$_p$      & 714  & 41.21 && 47.13  & 46.26  & 39.25  & 55.05  & 32.07  & 21.85  & 46.84 \\
&B$_p$      & 707  & 39.93 && 41.35  & 40.06  & 44.93  & 53.74  & 43.72  & 25.25  & 30.49 \\
&D          & 707  & 33.71 && 30.28  & 29.95  & 41.72  & 45.92  & 49.07  & 17.30  & 21.72 \\
&D$_p$      & 798  & 29.38 && 26.19  & 27.97  & 37.53  & 44.19  & 35.26  & 17.46  & 17.06 \\
&E          & 742  & 28.99 && 30.75  & 30.65  & 31.27  & 46.42  & 23.60  & 15.10  & 25.13 \\
&E$_p$      & 763  & 28.65 && 29.34  & 28.50  & 29.13  & 47.07  & 21.30  & 17.82  & 27.41 \\
\midrule


\multirow{10}*{\rotatebox{90}{Free Run 2}} 

&A          & 623  & 51.67 && 56.62  & 56.27  & 59.21  & 64.69  & 64.04  & 27.11  & 33.74 \\
&B          & 539  & 49.07 && 52.42  & 50.66  & 54.88  & 60.86  & 63.73  & 29.57  & 31.38 \\
&A$_p$      & 553  & 50.56 && 59.95  & 60.23  & 54.28  & 60.61  & 52.06  & 27.59  & 39.22 \\
&C          & 630  & 45.87 && 55.60  & 53.02  & 45.16  & 54.70  & 38.72  & 24.40  & 49.50 \\
&C$_p$      & 539  & 42.27 && 51.19  & 49.61  & 37.90  & 54.17  & 30.42  & 22.74  & 49.84 \\
&B$_p$      & 609  & 46.71 && 51.92  & 49.95  & 54.62  & 56.01  & 52.85  & 28.48  & 33.10 \\
&D          & 595  & 38.17 && 38.31  & 35.39  & 50.99  & 46.38  & 57.94  & 16.09  & 22.08 \\
&D$_p$      & 567  & 30.89 && 31.07  & 30.37  & 38.37  & 44.64  & 31.47  & 21.85  & 18.48 \\
&E          & 679  & 31.70 && 35.67  & 36.32  & 35.26  & 46.91  & 26.79  & 18.98  & 21.99 \\
&E$_p$      & 511  & 31.66 && 33.63  & 33.26  & 38.77  & 51.53  & 26.99  & 17.63  & 19.79 \\

\hline

\multicolumn{2}{c}{$r$} & $-$ &  0.959  && 0.947  & 0.919  & 0.880  & 0.951  & 0.951  & 0.783  & 0.945 \bigstrut[t]\\
\bottomrule
\end{tabular}%

%% file: tbl-04-system-raw-known.tex
\begin{tabular}{clcccccccccccc}
\toprule

&\rotatebox{60}{Model} & $n$ & \rotatebox{60}{Overall} && \rotatebox{60}{Interesting} & \rotatebox{60}{Fun} & \rotatebox{60}{Consistent} & \rotatebox{60}{Fluent} & \rotatebox{60}{Topic} & \rotatebox{60}{Robotic} & \rotatebox{60}{Repetitive} \bigstrut\\
\midrule

\multirow{10}*{\rotatebox{90}{Ice-breaker}} 
&A          & 721  & 53.43 && 53.65  & 52.35  & 63.24  & 67.28  & 66.97  & 28.17  & 42.32 \\
&A$_p$      & 721  & 50.21 && 54.53  & 53.50  & 52.84  & 58.83  & 53.18  & 38.87  & 39.70 \\
&B          & 742  & 49.55 && 49.23  & 47.76  & 57.79  & 60.64  & 62.22  & 32.56  & 36.65 \\
&C          & 784  & 47.93 && 56.18  & 53.69  & 43.15  & 56.88  & 40.46  & 29.61  & 55.54 \\
&B$_p$      & 700  & 44.94 && 48.83  & 46.70  & 49.58  & 55.86  & 49.21  & 25.82  & 38.61 \\
&C$_p$      & 658  & 42.41 && 47.98  & 45.48  & 37.66  & 54.51  & 32.50  & 26.00  & 52.72 \\
&D          & 728  & 35.14 && 30.32  & 33.13  & 42.90  & 49.92  & 48.51  & 20.11  & 21.09 \\
&E$_p$      & 721  & 31.58 && 31.73  & 30.82  & 35.44  & 47.12  & 27.06  & 21.90  & 26.97 \\
&E          & 721  & 30.09 && 33.17  & 31.95  & 31.14  & 47.12  & 24.90  & 19.10  & 23.23 \\
&D$_p$      & 714  & 27.22 && 22.56  & 22.53  & 35.22  & 41.70  & 34.98  & 17.44  & 16.09 \\

\midrule

\multicolumn{2}{c}{$r$} & $-$ & 0.970  && 0.955  & 0.918  & 0.949  & 0.928  & 0.972  & 0.738  & 0.968  \bigstrut[t]\\
\bottomrule
\end{tabular}%

%% file: fig-02-plots-chars.tex
\begin{tabular}{ccccccc}


\multirow{2}{*}{\rotatebox[origin=b]{90}{\parbox[c]{6cm}{\centering \footnotesize \# conversations }}} 
&& \\

&  \includegraphics[width=0.3\textwidth]{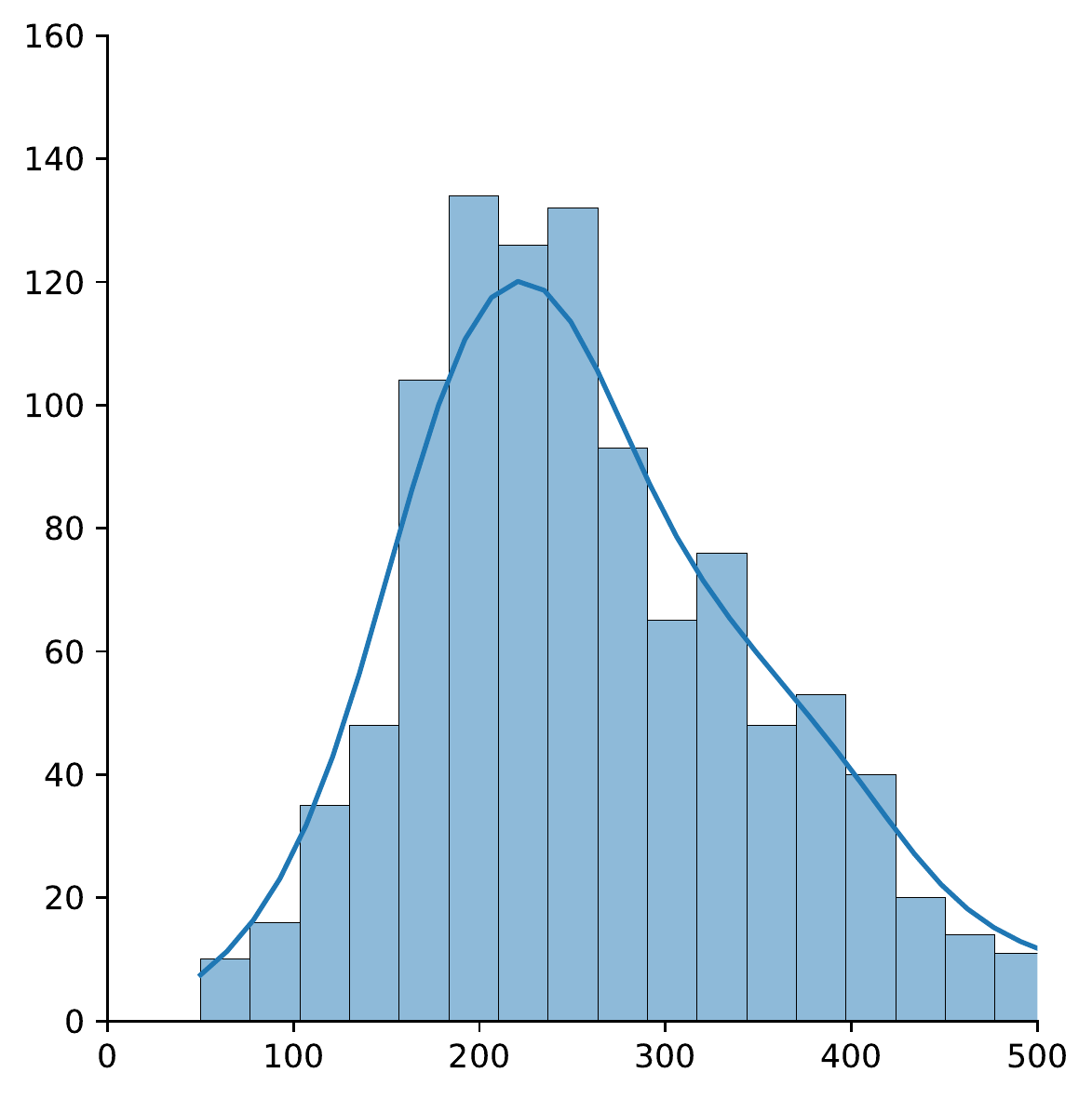}
& \includegraphics[width=0.3\textwidth]{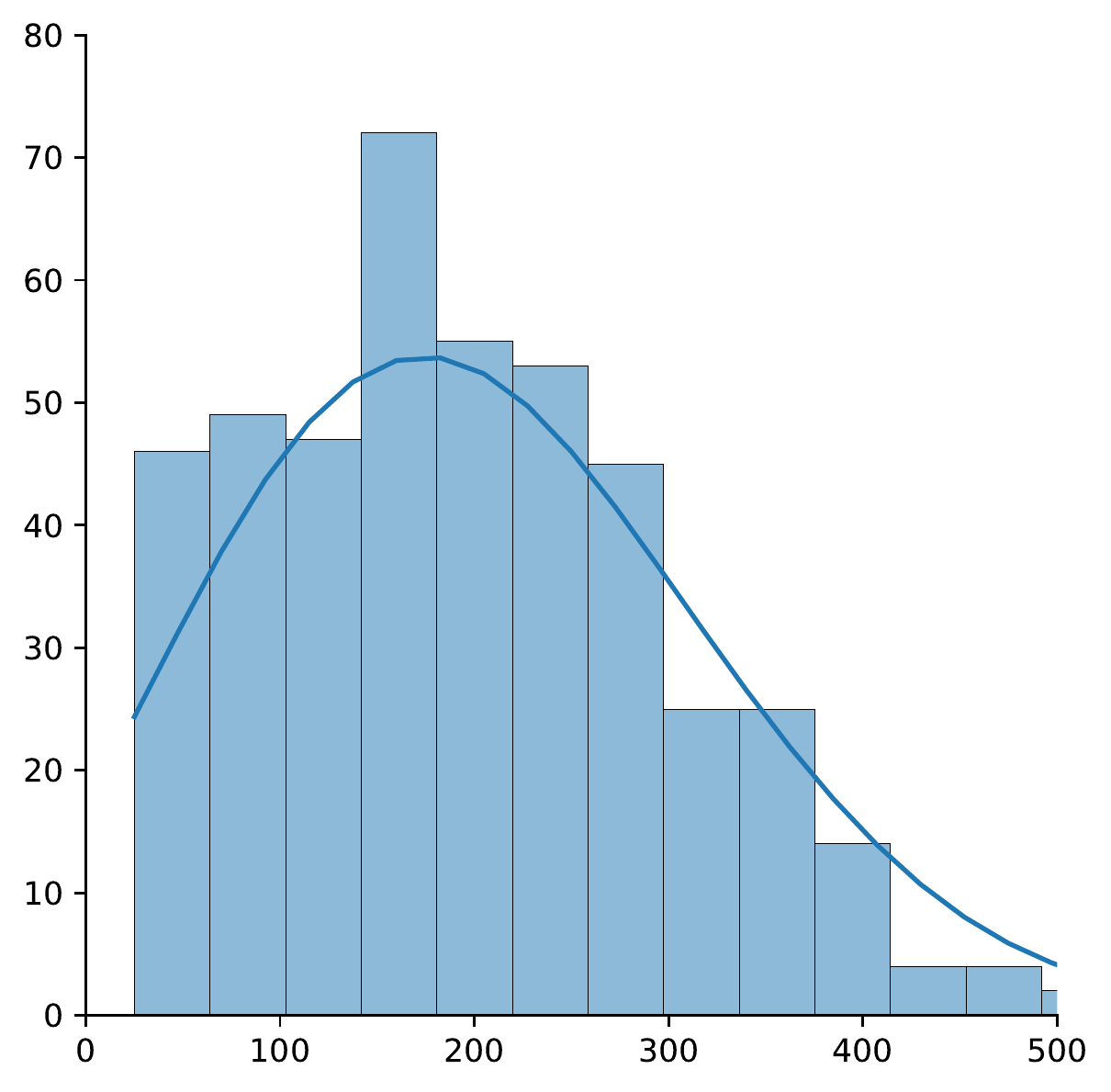} 
& \includegraphics[width=0.3\textwidth]{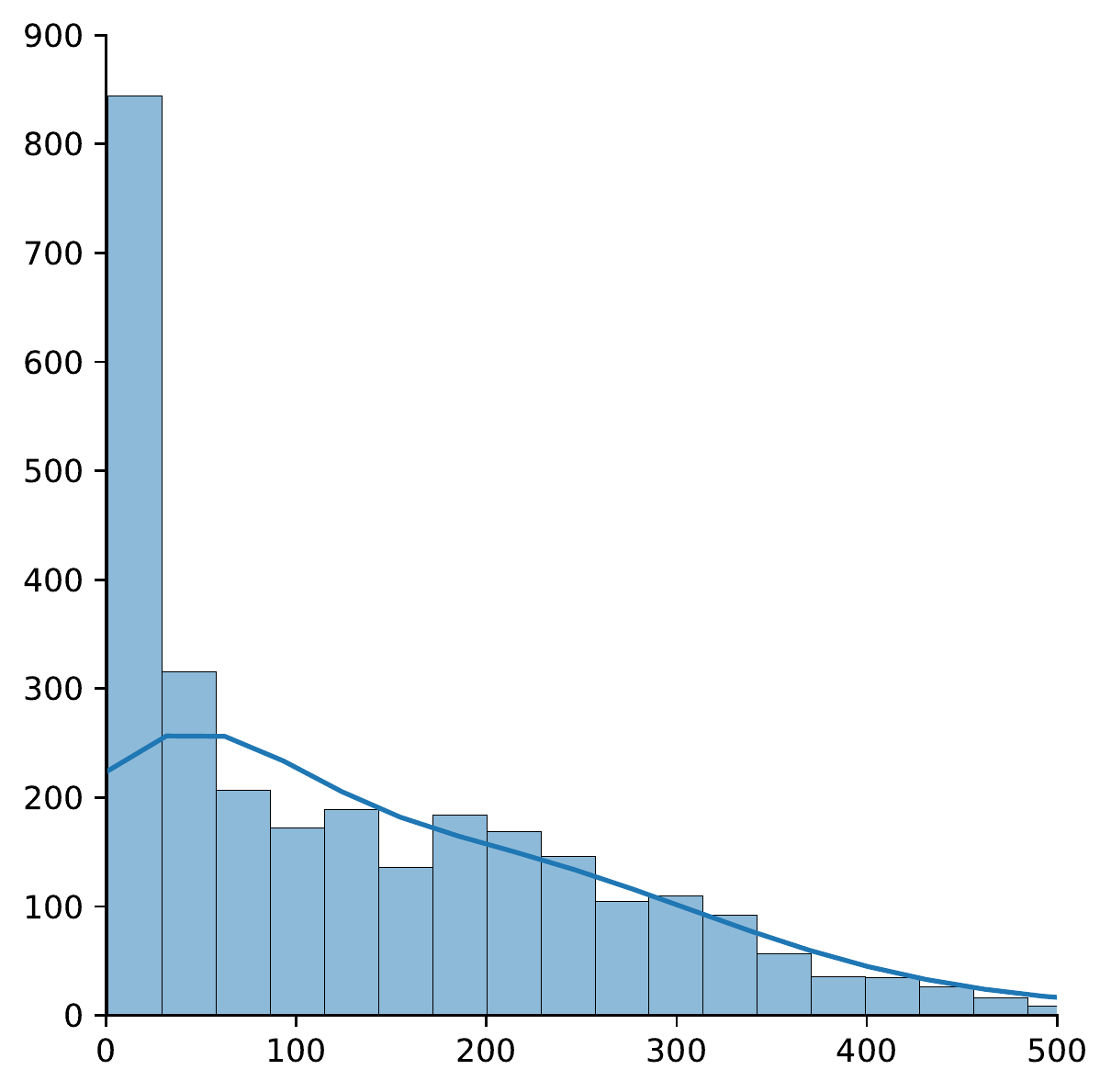}  \\
 
& \footnotesize \# characters & \footnotesize \# characters & \footnotesize \# characters \\[2ex]
     
&  \multicolumn{1}{c}{(a) Pass Quality Control} &  \multicolumn{1}{c}{(b) Fail Quality Control} &  \multicolumn{1}{c}{(c) ConvAI2 Live}\\

\end{tabular}

%% file: fig-03-plots-words.tex
\begin{tabular}{ccccccc}

\multirow{2}{*}{\rotatebox[origin=b]{90}{\parbox[c]{6cm}{\centering \footnotesize \# conversations }}} 
&& \\

&  \includegraphics[width=0.3\textwidth]{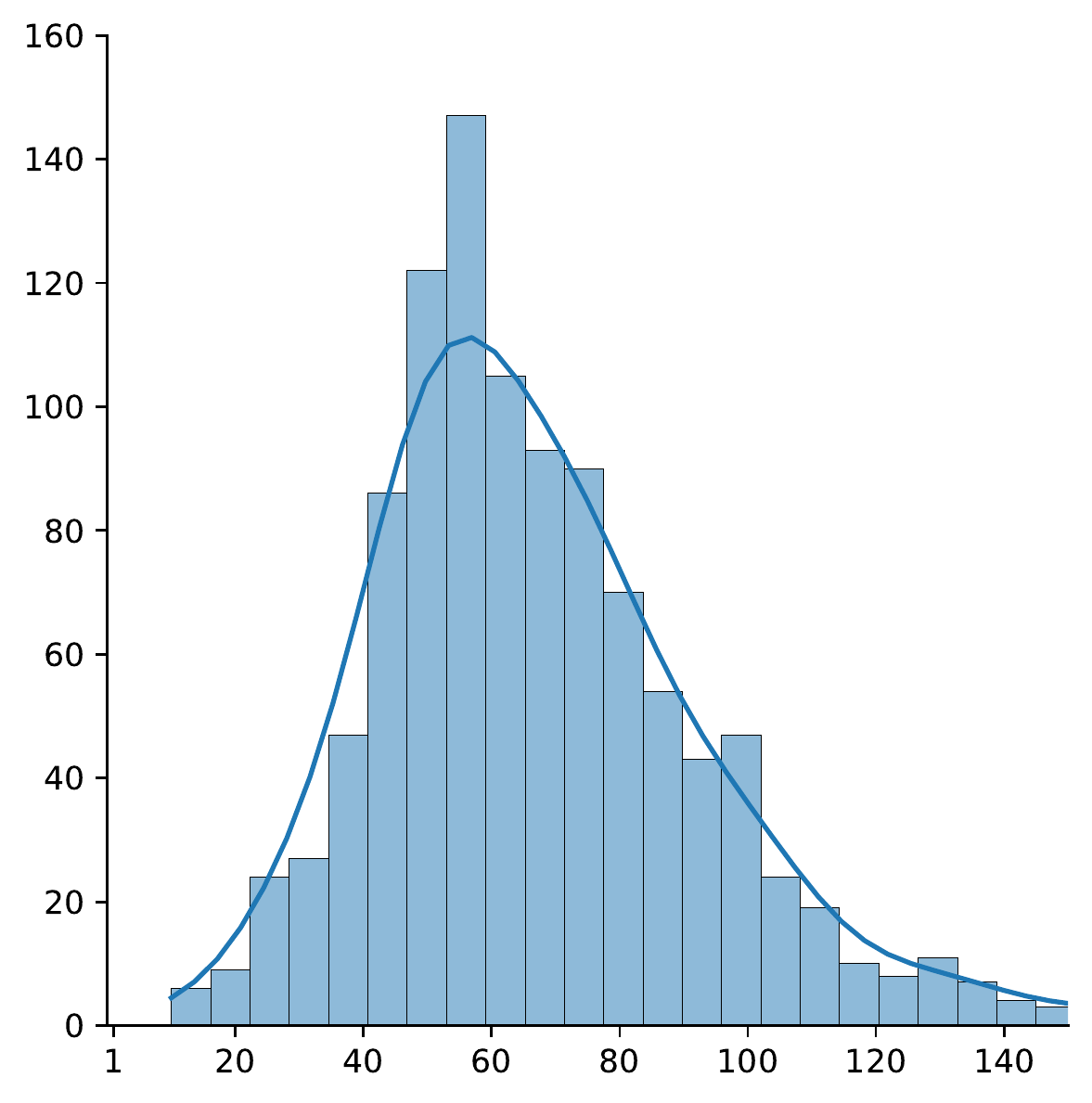}
& \includegraphics[width=0.3\textwidth]{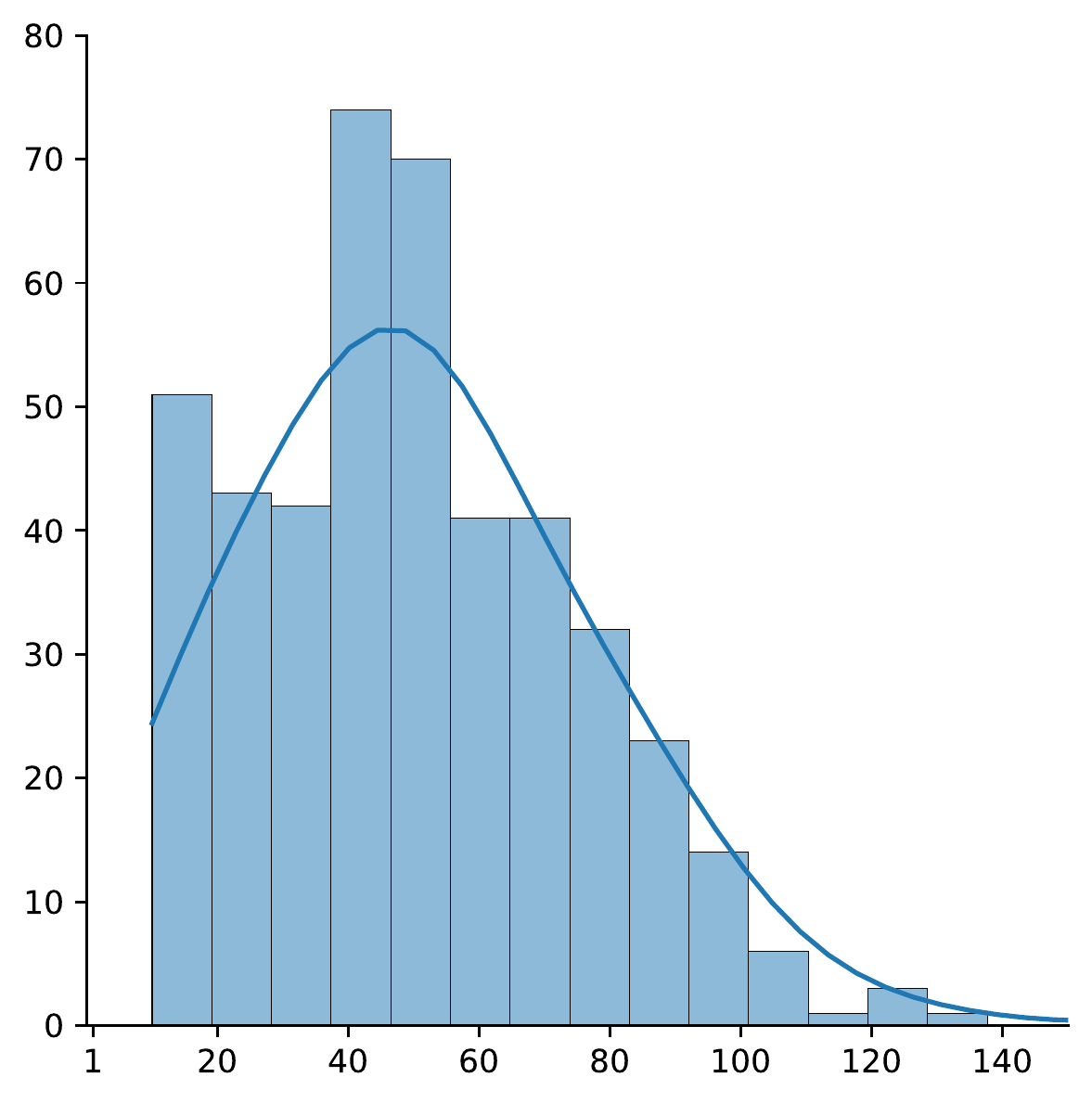}
& \includegraphics[width=0.3\textwidth]{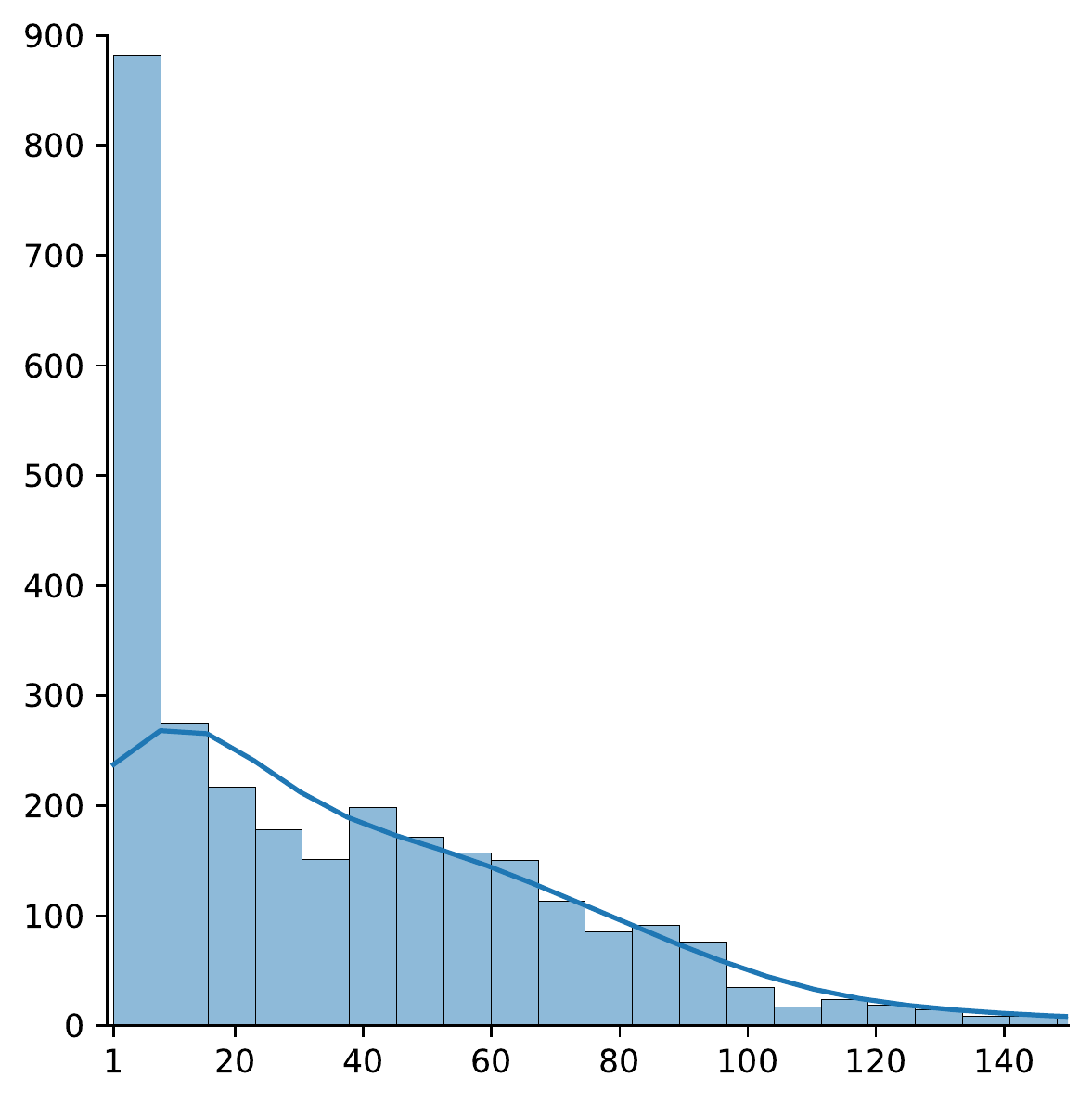}  \\
 
& \footnotesize \# words & \footnotesize \# words & \footnotesize \# words \\[2ex]
     
&  \multicolumn{1}{c}{(a) Pass Quality Control} &  \multicolumn{1}{c}{(b) Fail Quality Control} &  \multicolumn{1}{c}{(c) ConvAI2 Live}\\

\end{tabular}

%% file: fig-04-plots-chars-uttr.tex
\begin{tabular}{ccccccc}


\multirow{2}{*}{\rotatebox[origin=b]{90}{\parbox[c]{6cm}{\centering \footnotesize \# inputs }}} 
&& \\

&  \includegraphics[width=0.3\textwidth]{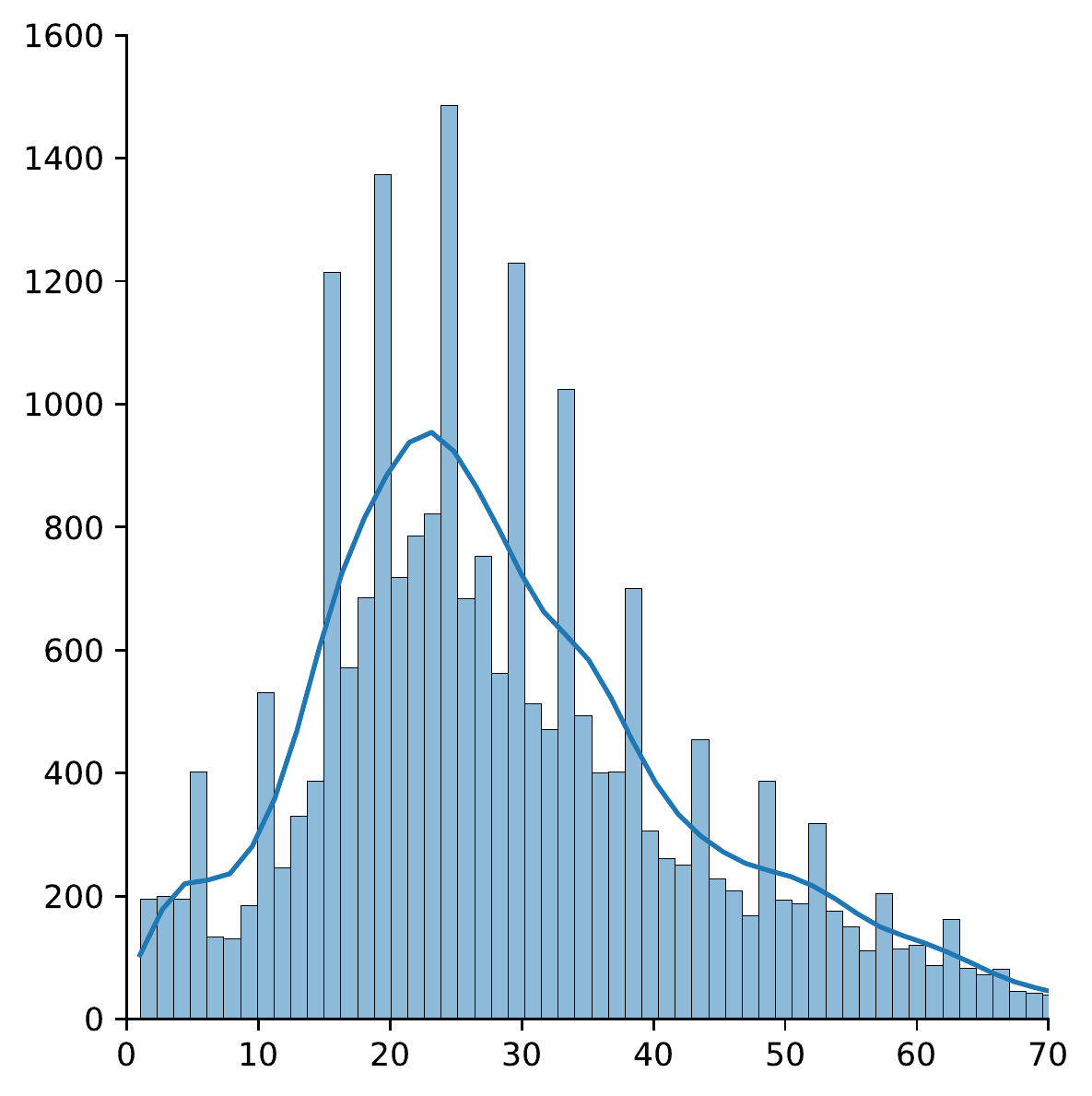}
& \includegraphics[width=0.3\textwidth]{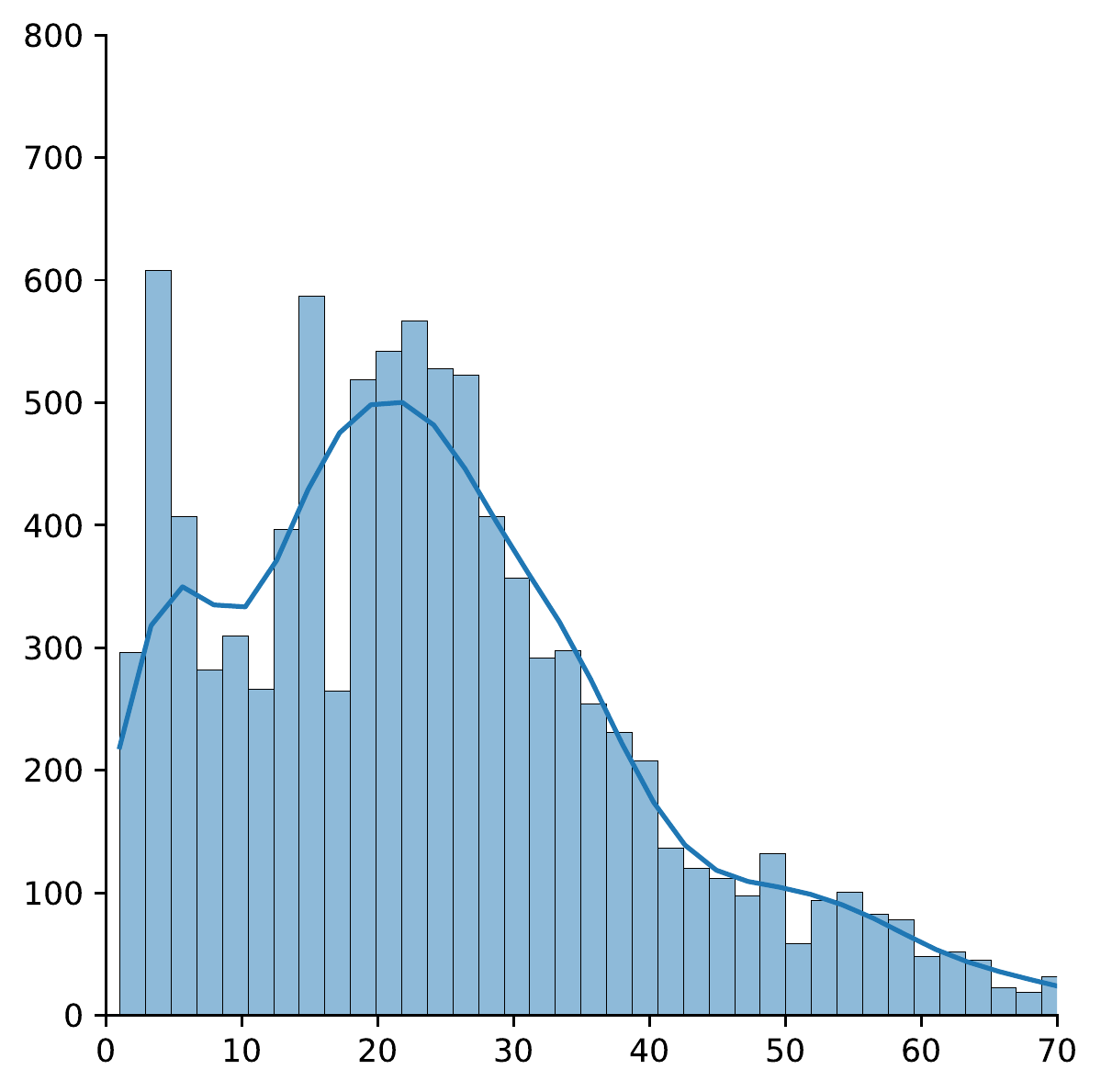}
& \includegraphics[width=0.3\textwidth]{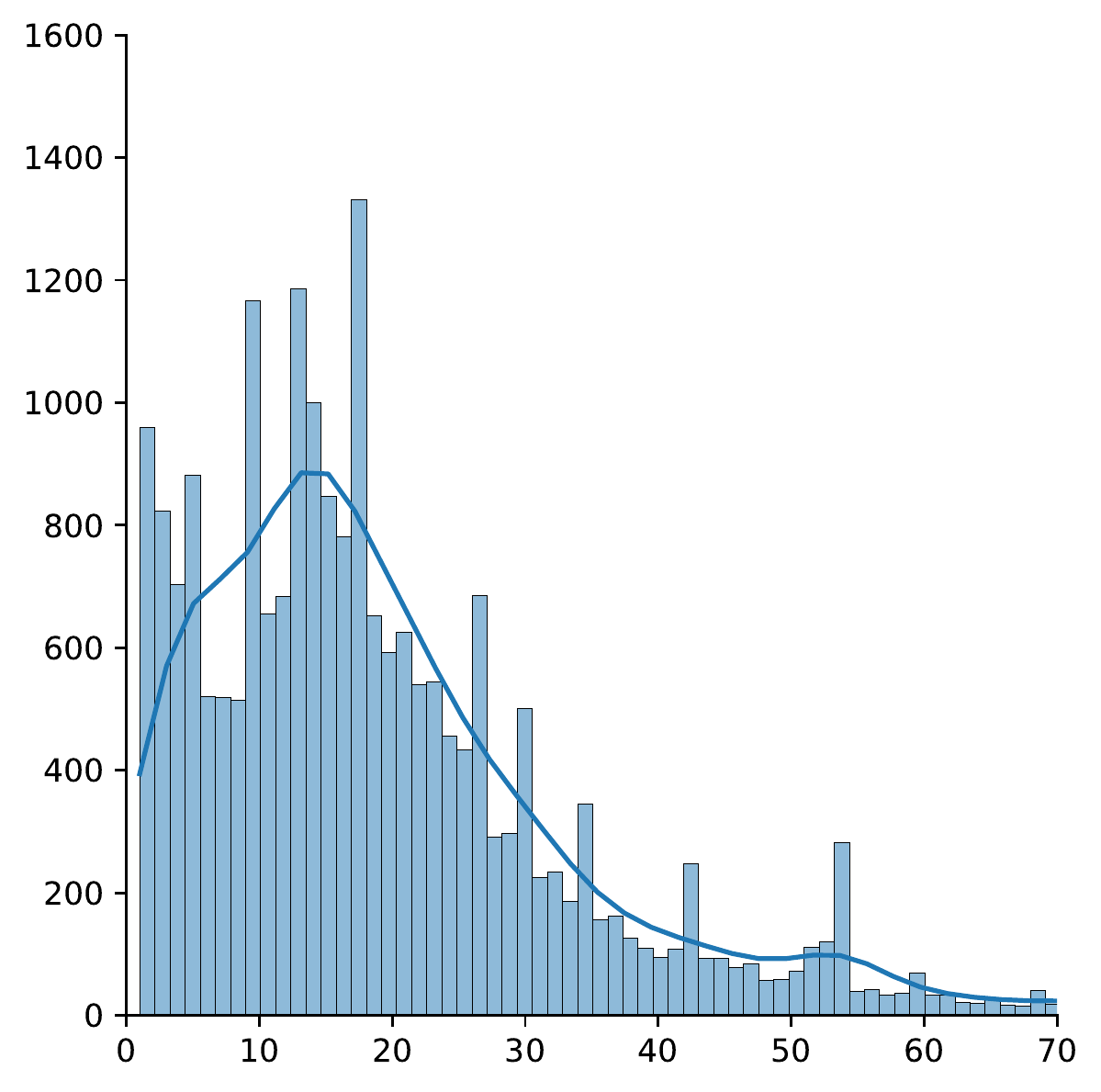}  \\
 
& \footnotesize \# characters & \footnotesize \# characters & \footnotesize \# characters \\[2ex]
     
&  \multicolumn{1}{c}{(a) Pass Quality Control} &  \multicolumn{1}{c}{(b) Fail Quality Control} &  \multicolumn{1}{c}{(c) ConvAI2 Live}\\

\end{tabular}

%% file: fig-05-plots-words-uttr.tex
\begin{tabular}{ccccccc}

\multirow{2}{*}{\rotatebox[origin=b]{90}{\parbox[c]{6cm}{\centering \footnotesize \# inputs }}} 
&& \\


&  \includegraphics[width=0.3\textwidth]{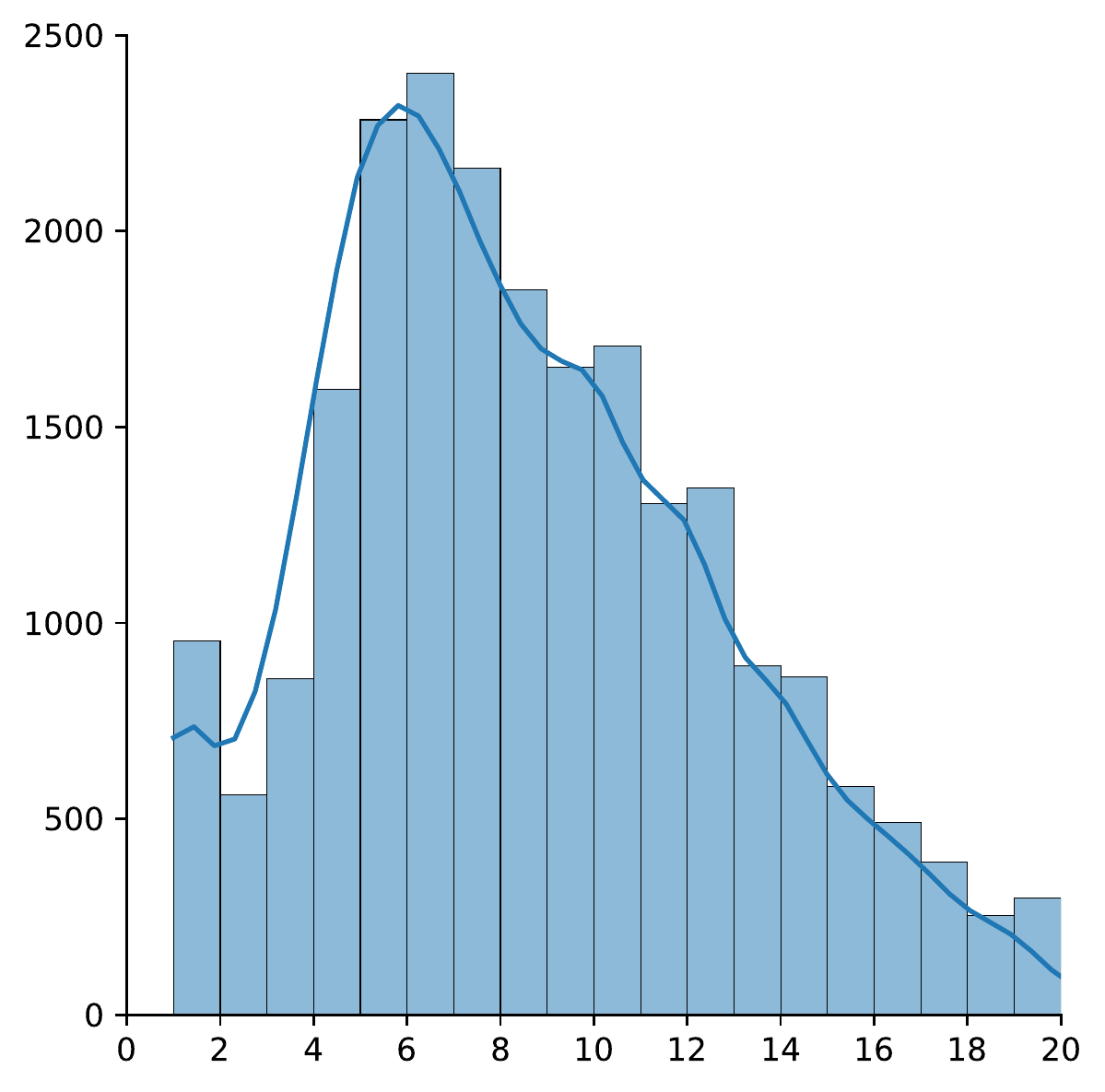}
& \includegraphics[width=0.3\textwidth]{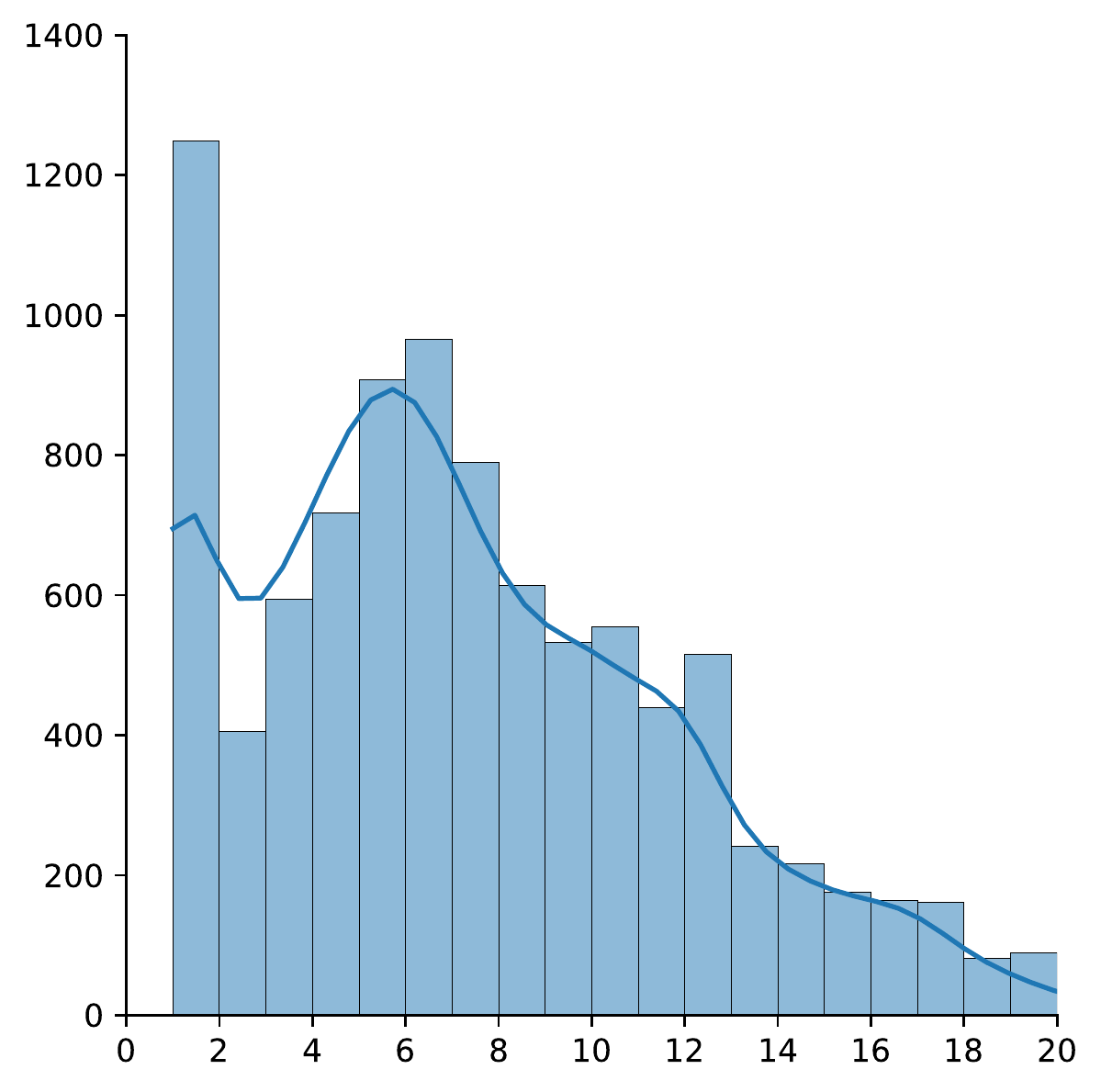}
& \includegraphics[width=0.3\textwidth]{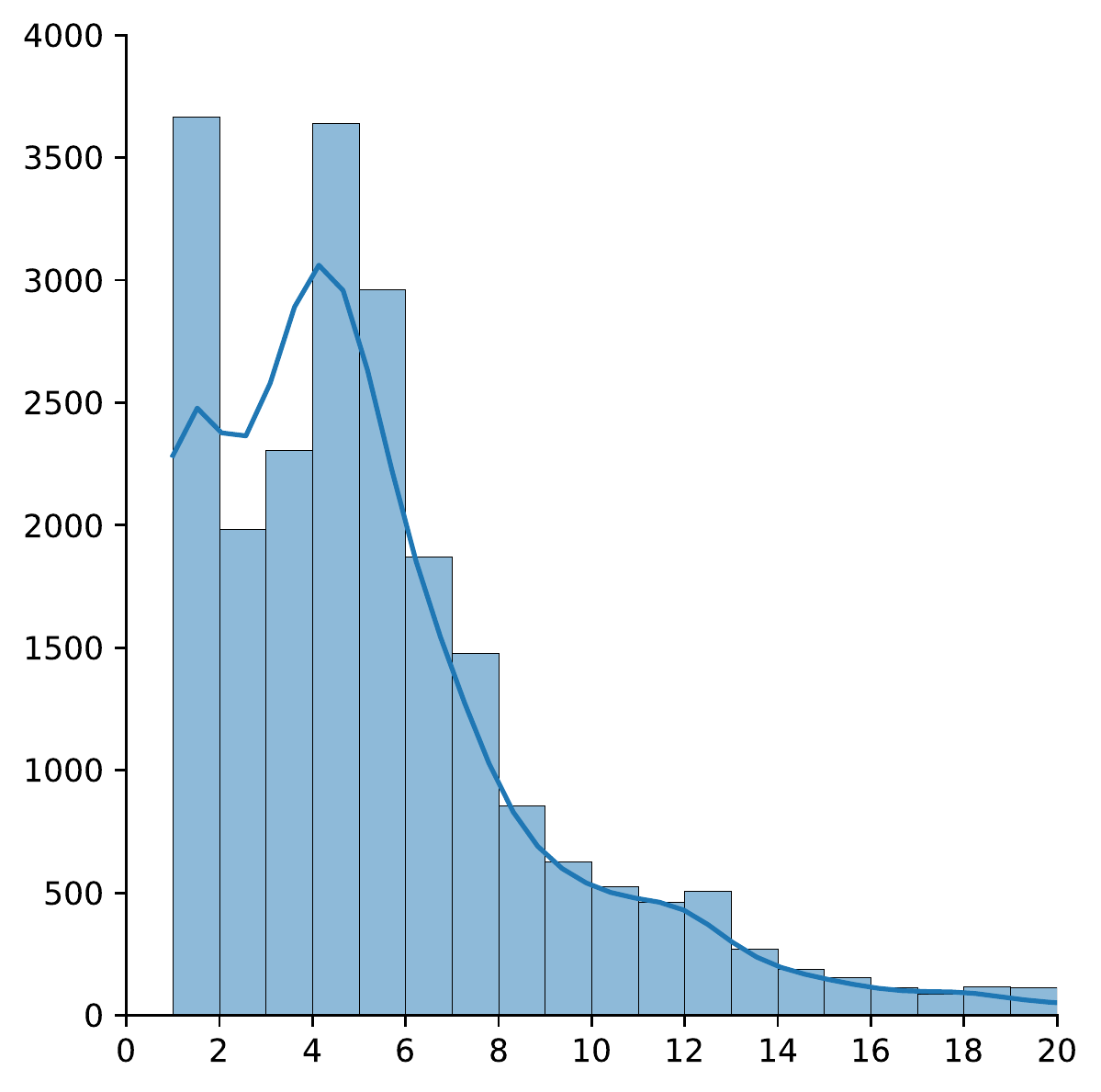}  \\
 
& \footnotesize \# words & \footnotesize \# words & \footnotesize \# words \\[2ex]
     
&  \multicolumn{1}{c}{(a) Pass Quality Control} &  \multicolumn{1}{c}{(b) Fail Quality Control} &  \multicolumn{1}{c}{(c) ConvAI2 Live}\\

\end{tabular}

%% file: tbl-09-words-chars.tex
\begin{tabular}{p{2.5cm}lrrr}
\toprule
                                && \multicolumn{1}{c}{Passed QC} & \multicolumn{1}{c}{Failed QC} & \multicolumn{1}{c}{ConvAI2} \\
\midrule
\multirow{2}{*}{Characters} & Median in an Input        & 27 & 22 & 16 \\
& Median in a Conversation  & 249 & 188 & 105 \\[2ex]

\multirow{2}{*}{Words} & Median in an Input             &  8 &  6 &  4 \\
& Median in a Conversation       & 63 & 48 & 28 \\

\bottomrule

\end{tabular}

%% file: acl2022.bbl
\begin{thebibliography}{35}
\expandafter\ifx\csname natexlab\endcsname\relax\def\natexlab#1{#1}\fi

\bibitem[{Alexander(1990)}]{alexander1990note}
Ralph~A Alexander. 1990.
\newblock A note on averaging correlations.
\newblock \emph{Bulletin of the Psychonomic Society}, 28(4):335--336.

\bibitem[{Alexandrov(2010)}]{alexandrov2010characteristics}
Aliosha Alexandrov. 2010.
\newblock Characteristics of single-item measures in likert scale format.
\newblock \emph{The Electronic Journal of Business Research Methods},
  8(1):1--12.

\bibitem[{Barrault et~al.(2020)Barrault, Biesialska, Bojar, Costa-jussà,
  Federmann, Graham, Grundkiewicz, Haddow, Huck, Joanis, Kocmi, Koehn, Lo,
  Ljubešić, Monz, Morishita, Nagata, Nakazawa, Pal, Post, and
  Zampieri}]{wmt20}
Loïc Barrault, Magdalena Biesialska, Ondřej Bojar, Marta~R. Costa-jussà,
  Christian Federmann, Yvette Graham, Roman Grundkiewicz, Barry Haddow,
  Matthias Huck, Eric Joanis, Tom Kocmi, Philipp Koehn, Chi-kiu Lo, Nikola
  Ljubešić, Christof Monz, Makoto Morishita, Masaaki Nagata, Toshiaki
  Nakazawa, Santanu Pal, Matt Post, and Marcos Zampieri. 2020.
\newblock \href {https://www.aclweb.org/anthology/2020.wmt-1.1} {Findings of
  the 2020 conference on machine translation (wmt20)}.
\newblock In \emph{Proceedings of the Fifth Conference on Machine Translation},
  pages 1--54, Online. Association for Computational Linguistics.

\bibitem[{Bojar et~al.(2014)Bojar, Buck, Federmann, Haddow, Koehn, Leveling,
  Monz, Pecina, Post, Saint-Amand, Soricut, Specia, and Tamchyna}]{wmt14}
Ondrej Bojar, Christian Buck, Christian Federmann, Barry Haddow, Philipp Koehn,
  Johannes Leveling, Christof Monz, Pavel Pecina, Matt Post, Herve Saint-Amand,
  Radu Soricut, Lucia Specia, and Ale\v{s} Tamchyna. 2014.
\newblock \href {http://www.aclweb.org/anthology/W/W14/W14-3302} {Findings of
  the 2014 workshop on statistical machine translation}.
\newblock In \emph{Proceedings of the Ninth Workshop on Statistical Machine
  Translation}, pages 12--58, Baltimore, Maryland, USA. Association for
  Computational Linguistics.

\bibitem[{Bojar et~al.(2013)Bojar, Buck, Callison-Burch, Federmann, Haddow,
  Koehn, Monz, Post, Soricut, and Specia}]{wmt13}
Ond\v{r}ej Bojar, Christian Buck, Chris Callison-Burch, Christian Federmann,
  Barry Haddow, Philipp Koehn, Christof Monz, Matt Post, Radu Soricut, and
  Lucia Specia. 2013.
\newblock \href {http://www.aclweb.org/anthology/W13-2201} {Findings of the
  2013 {Workshop on Statistical Machine Translation}}.
\newblock In \emph{Proceedings of the Eighth Workshop on Statistical Machine
  Translation}, pages 1--44, Sofia, Bulgaria. Association for Computational
  Linguistics.

\bibitem[{Callison-Burch et~al.(2012)Callison-Burch, Koehn, Monz, Post,
  Soricut, and Specia}]{wmt12}
Chris Callison-Burch, Philipp Koehn, Christof Monz, Matt Post, Radu Soricut,
  and Lucia Specia. 2012.
\newblock \href {http://www.aclweb.org/anthology/W12-3102} {Findings of the
  2012 workshop on statistical machine translation}.
\newblock In \emph{Proceedings of the Seventh Workshop on Statistical Machine
  Translation}, pages 10--51, Montr{\'e}al, Canada. Association for
  Computational Linguistics.

\bibitem[{Callison-Burch et~al.(2011)Callison-Burch, Koehn, Monz, and
  Zaidan}]{wmt11}
Chris Callison-Burch, Philipp Koehn, Christof Monz, and Omar Zaidan. 2011.
\newblock \href {http://www.aclweb.org/anthology/W11-2103} {Findings of the
  2011 workshop on statistical machine translation}.
\newblock In \emph{Proceedings of the Sixth Workshop on Statistical Machine
  Translation}, pages 22--64, Edinburgh, Scotland. Association for
  Computational Linguistics.

\bibitem[{Denkowski and Lavie(2011)}]{metrics-meteor}
M.~Denkowski and A.~Lavie. 2011.
\newblock Meteor 1.3: Automatic metric for reliable optimization and evaluation
  of machine translation systems.
\newblock In \emph{Proceedings of the Sixth Workshop on Statistical Machine
  Translation}, pages 85--91. Association for Computational Linguistics.

\bibitem[{Dinan et~al.(2019)Dinan, Logacheva, Malykh, Miller, Shuster, Urbanek,
  Kiela, Szlam, Serban, Lowe, Prabhumoye, Black, Rudnicky, Williams, Pineau,
  Burtsev, and Weston}]{convai2paper}
Emily Dinan, Varvara Logacheva, Valentin Malykh, Alexander~H. Miller, Kurt
  Shuster, Jack Urbanek, Douwe Kiela, Arthur Szlam, Iulian Serban, Ryan Lowe,
  Shrimai Prabhumoye, Alan~W. Black, Alexander~I. Rudnicky, Jason Williams,
  Joelle Pineau, Mikhail Burtsev, and Jason Weston. 2019.
\newblock \href {http://arxiv.org/abs/1902.00098} {The second conversational
  intelligence challenge (convai2)}.
\newblock \emph{CoRR}, abs/1902.00098.

\bibitem[{Dinan et~al.(2018)Dinan, Roller, Shuster, Fan, Auli, and
  Weston}]{biencoder2}
Emily Dinan, Stephen Roller, Kurt Shuster, Angela Fan, Michael Auli, and Jason
  Weston. 2018.
\newblock \href {http://arxiv.org/abs/1811.01241} {Wizard of wikipedia:
  Knowledge-powered conversational agents}.
\newblock \emph{CoRR}, abs/1811.01241.

\bibitem[{Finch and Choi(2020)}]{current-evaluation-dialogue}
Sarah~E. Finch and Jinho~D. Choi. 2020.
\newblock \href {https://www.aclweb.org/anthology/2020.sigdial-1.29} {Towards
  unified dialogue system evaluation: A comprehensive analysis of current
  evaluation protocols}.
\newblock In \emph{Proceedings of the 21th Annual Meeting of the Special
  Interest Group on Discourse and Dialogue}, pages 236--245, 1st virtual
  meeting. Association for Computational Linguistics.

\bibitem[{Graham et~al.(2013)Graham, Baldwin, Moffat, and
  Zobel}]{graham-etal-2013-crowd}
Yvette Graham, Timothy Baldwin, Alistair Moffat, and Justin Zobel. 2013.
\newblock \href {https://www.aclweb.org/anthology/U13-1004} {Crowd-sourcing of
  human judgments of machine translation fluency}.
\newblock In \emph{Proceedings of the Australasian Language Technology
  Association Workshop 2013 ({ALTA} 2013)}, pages 16--24, Brisbane, Australia.

\bibitem[{Hochreiter and Schmidhuber(1997)}]{lstm-based}
Sepp Hochreiter and J\"{u}rgen Schmidhuber. 1997.
\newblock \href {https://doi.org/10.1162/neco.1997.9.8.1735} {Long short-term
  memory}.
\newblock \emph{Neural Comput.}, 9(8):1735–1780.

\bibitem[{Howcroft et~al.(2020)Howcroft, Belz, Clinciu, Gkatzia, Hasan,
  Mahamood, Mille, van Miltenburg, Santhanam, and
  Rieser}]{howcroft-etal-2020-twenty_years}
David~M. Howcroft, Anya Belz, Miruna-Adriana Clinciu, Dimitra Gkatzia, Sadid~A.
  Hasan, Saad Mahamood, Simon Mille, Emiel van Miltenburg, Sashank Santhanam,
  and Verena Rieser. 2020.
\newblock \href {https://aclanthology.org/2020.inlg-1.23} {Twenty years of
  confusion in human evaluation: {NLG} needs evaluation sheets and standardised
  definitions}.
\newblock In \emph{Proceedings of the 13th International Conference on Natural
  Language Generation}, pages 169--182, Dublin, Ireland. Association for
  Computational Linguistics.

\bibitem[{Humeau et~al.(2019)Humeau, Shuster, Lachaux, and
  Weston}]{polyencoder}
Samuel Humeau, Kurt Shuster, Marie{-}Anne Lachaux, and Jason Weston. 2019.
\newblock \href {http://arxiv.org/abs/1905.01969} {Poly{-}encoders: Transformer
  architectures and pre-training strategies for fast and accurate
  multi-sentence scoring}.
\newblock \emph{CoRR}, abs/1905.01969.

\bibitem[{Lasecki et~al.(2014)Lasecki, Teevan, and Kamar}]{gold-standard-03}
Walter~S. Lasecki, Jaime Teevan, and Ece Kamar. 2014.
\newblock \href {https://doi.org/10.1145/2531602.2531733} {Information
  extraction and manipulation threats in crowd-powered systems}.
\newblock In \emph{Proceedings of the 17th ACM Conference on Computer Supported
  Cooperative Work \& Social Computing}, CSCW '14, page 248–256, New York,
  NY, USA. Association for Computing Machinery.

\bibitem[{Li et~al.(2019)Li, Weston, and Roller}]{li2019acute}
Margaret Li, Jason Weston, and Stephen Roller. 2019.
\newblock Acute-eval: Improved dialogue evaluation with optimized questions and
  multi-turn comparisons.
\newblock \emph{arXiv preprint arXiv:1909.03087}.

\bibitem[{Lin and Hovy(2003)}]{metrics-rouge}
Chin-Yew Lin and Eduard Hovy. 2003.
\newblock Automatic evaluation of summaries using n-gram co-occurrence
  statistics.
\newblock In \emph{Proceedings of the 2003 Conference of the North American
  Chapter of the Association for Computational Linguistics on Human Language
  Technology-Volume 1}, pages 71--78. Association for Computational
  Linguistics.

\bibitem[{Liu et~al.(2013)Liu, Ihler, and Steyvers}]{gold-standard-02}
Qiang Liu, Alexander~T Ihler, and Mark Steyvers. 2013.
\newblock \href
  {https://proceedings.neurips.cc/paper/2013/file/cc1aa436277138f61cda703991069eaf-Paper.pdf}
  {Scoring workers in crowdsourcing: How many control questions are enough?}
\newblock In \emph{Advances in Neural Information Processing Systems},
  volume~26. Curran Associates, Inc.

\bibitem[{Liu et~al.(2019)Liu, Ott, Goyal, Du, Joshi, Chen, Levy, Lewis,
  Zettlemoyer, and Stoyanov}]{model-roberta}
Yinhan Liu, Myle Ott, Naman Goyal, Jingfei Du, Mandar Joshi, Danqi Chen, Omer
  Levy, Mike Lewis, Luke Zettlemoyer, and Veselin Stoyanov. 2019.
\newblock \href {http://arxiv.org/abs/1907.11692} {Roberta: {A} robustly
  optimized {BERT} pretraining approach}.
\newblock \emph{CoRR}, abs/1907.11692.

\bibitem[{Loukina et~al.(2020)Loukina, Madnani, Cahill, Yao, Johnson, Riordan,
  and McCaffrey}]{loukina-etal-2020-using}
Anastassia Loukina, Nitin Madnani, Aoife Cahill, Lili Yao, Matthew~S. Johnson,
  Brian Riordan, and Daniel~F. McCaffrey. 2020.
\newblock \href {https://doi.org/10.18653/v1/2020.bea-1.2} {Using {PRMSE} to
  evaluate automated scoring systems in the presence of label noise}.
\newblock In \emph{Proceedings of the Fifteenth Workshop on Innovative Use of
  NLP for Building Educational Applications}, pages 18--29, Seattle, WA, USA
  → Online. Association for Computational Linguistics.

\bibitem[{Mehri and
  Eskenazi(2020{\natexlab{a}})}]{metric-unsupervised-evaluation-dialogpt}
Shikib Mehri and Maxine Eskenazi. 2020{\natexlab{a}}.
\newblock \href {https://www.aclweb.org/anthology/2020.sigdial-1.28}
  {Unsupervised evaluation of interactive dialog with {D}ialo{GPT}}.
\newblock In \emph{Proceedings of the 21th Annual Meeting of the Special
  Interest Group on Discourse and Dialogue}, pages 225--235, 1st virtual
  meeting. Association for Computational Linguistics.

\bibitem[{Mehri and Eskenazi(2020{\natexlab{b}})}]{metric-usr}
Shikib Mehri and Maxine Eskenazi. 2020{\natexlab{b}}.
\newblock \href {https://doi.org/10.18653/v1/2020.acl-main.64} {{USR}: An
  unsupervised and reference free evaluation metric for dialog generation}.
\newblock In \emph{Proceedings of the 58th Annual Meeting of the Association
  for Computational Linguistics}, pages 681--707, Online. Association for
  Computational Linguistics.

\bibitem[{Mille et~al.(2020)Mille, Belz, Bohnet, Castro~Ferreira, Graham, and
  Wanner}]{mille-etal-2020-third}
Simon Mille, Anya Belz, Bernd Bohnet, Thiago Castro~Ferreira, Yvette Graham,
  and Leo Wanner. 2020.
\newblock \href {https://www.aclweb.org/anthology/2020.msr-1.1} {The third
  multilingual surface realisation shared task ({SR}{'}20): Overview and
  evaluation results}.
\newblock In \emph{Proceedings of the Third Workshop on Multilingual Surface
  Realisation}, pages 1--20, Barcelona, Spain (Online). Association for
  Computational Linguistics.

\bibitem[{Miller et~al.(2016)Miller, Fisch, Dodge, Karimi, Bordes, and
  Weston}]{kvmn}
Alexander~H. Miller, Adam Fisch, Jesse Dodge, Amir{-}Hossein Karimi, Antoine
  Bordes, and Jason Weston. 2016.
\newblock \href {http://arxiv.org/abs/1606.03126} {Key-value memory networks
  for directly reading documents}.
\newblock \emph{CoRR}, abs/1606.03126.

\bibitem[{Novikova et~al.(2018)Novikova, Du{\v{s}}ek, and
  Rieser}]{novikova-etal-2018-rankme}
Jekaterina Novikova, Ond{\v{r}}ej Du{\v{s}}ek, and Verena Rieser. 2018.
\newblock \href {https://doi.org/10.18653/v1/N18-2012} {{R}ank{ME}: Reliable
  human ratings for natural language generation}.
\newblock In \emph{Proceedings of the 2018 Conference of the North {A}merican
  Chapter of the Association for Computational Linguistics: Human Language
  Technologies, Volume 2 (Short Papers)}, pages 72--78, New Orleans, Louisiana.
  Association for Computational Linguistics.

\bibitem[{Pang et~al.(2020)Pang, Nijkamp, Han, Zhou, Liu, and
  Tu}]{metric-towards-holistic}
Bo~Pang, Erik Nijkamp, Wenjuan Han, Linqi Zhou, Yixian Liu, and Kewei Tu. 2020.
\newblock \href {https://doi.org/10.18653/v1/2020.acl-main.333} {Towards
  holistic and automatic evaluation of open-domain dialogue generation}.
\newblock In \emph{Proceedings of the 58th Annual Meeting of the Association
  for Computational Linguistics}, pages 3619--3629, Online. Association for
  Computational Linguistics.

\bibitem[{Papineni et~al.(2002)Papineni, Roukos, Ward, and Zhu}]{metrics-bleu}
Kishore Papineni, Salim Roukos, Todd Ward, and Wei-Jing Zhu. 2002.
\newblock \href {https://doi.org/10.3115/1073083.1073135} {Bleu: A method for
  automatic evaluation of machine translation}.
\newblock In \emph{Proceedings of the 40th Annual Meeting on Association for
  Computational Linguistics}, ACL '02, pages 311--318, Stroudsburg, PA, USA.
  Association for Computational Linguistics.

\bibitem[{Ram et~al.(2018)Ram, Prasad, Khatri, Venkatesh, Gabriel, Liu, Nunn,
  Hedayatnia, Cheng, Nagar, King, Bland, Wartick, Pan, Song, Jayadevan, Hwang,
  and Pettigrue}]{alexa1}
Ashwin Ram, Rohit Prasad, Chandra Khatri, Anu Venkatesh, Raefer Gabriel, Qing
  Liu, Jeff Nunn, Behnam Hedayatnia, Ming Cheng, Ashish Nagar, Eric King, Kate
  Bland, Amanda Wartick, Yi~Pan, Han Song, Sk~Jayadevan, Gene Hwang, and Art
  Pettigrue. 2018.
\newblock \href {http://arxiv.org/abs/1801.03604} {Conversational {AI:} the
  science behind the alexa prize}.
\newblock \emph{CoRR}, abs/1801.03604.

\bibitem[{Santhanam et~al.(2020)Santhanam, Karduni, and
  Shaikh}]{santhanam2020studying}
Sashank Santhanam, Alireza Karduni, and Samira Shaikh. 2020.
\newblock Studying the effects of cognitive biases in evaluation of
  conversational agents.
\newblock In \emph{Proceedings of the 2020 CHI Conference on Human Factors in
  Computing Systems}, pages 1--13.

\bibitem[{Santhanam and Shaikh(2019)}]{santhanam-shaikh-2019-towards}
Sashank Santhanam and Samira Shaikh. 2019.
\newblock \href {https://doi.org/10.18653/v1/W19-8610} {Towards best experiment
  design for evaluating dialogue system output}.
\newblock In \emph{Proceedings of the 12th International Conference on Natural
  Language Generation}, pages 88--94, Tokyo, Japan. Association for
  Computational Linguistics.

\bibitem[{Sorodoc et~al.(2017)Sorodoc, Lau, Aletras, and
  Baldwin}]{sorodoc-etal-2017-multimodal}
Ionut Sorodoc, Jey~Han Lau, Nikolaos Aletras, and Timothy Baldwin. 2017.
\newblock \href {https://www.aclweb.org/anthology/E17-2111} {Multimodal topic
  labelling}.
\newblock In \emph{Proceedings of the 15th Conference of the {E}uropean Chapter
  of the Association for Computational Linguistics: Volume 2, Short Papers},
  pages 701--706, Valencia, Spain. Association for Computational Linguistics.

\bibitem[{Sutskever et~al.(2014)Sutskever, Vinyals, and Le}]{seq2seq}
Ilya Sutskever, Oriol Vinyals, and Quoc~V. Le. 2014.
\newblock Sequence to sequence learning with neural networks.
\newblock In \emph{Proceedings of the 27th International Conference on Neural
  Information Processing Systems - Volume 2}, NIPS'14, page 3104–3112,
  Cambridge, MA, USA. MIT Press.

\bibitem[{Wu et~al.(2016)Wu, Schuster, Chen, Le, Norouzi, Macherey, Krikun,
  Cao, Gao, Macherey, Klingner, Shah, Johnson, Liu, Kaiser, Gouws, Kato, Kudo,
  Kazawa, Stevens, Kurian, Patil, Wang, Young, Smith, Riesa, Rudnick, Vinyals,
  Corrado, Hughes, and Dean}]{metrics-gleu}
Yonghui Wu, Mike Schuster, Zhifeng Chen, Quoc~V. Le, Mohammad Norouzi, Wolfgang
  Macherey, Maxim Krikun, Yuan Cao, Qin Gao, Klaus Macherey, Jeff Klingner,
  Apurva Shah, Melvin Johnson, Xiaobing Liu, Lukasz Kaiser, Stephan Gouws,
  Yoshikiyo Kato, Taku Kudo, Hideto Kazawa, Keith Stevens, George Kurian,
  Nishant Patil, Wei Wang, Cliff Young, Jason Smith, Jason Riesa, Alex Rudnick,
  Oriol Vinyals, Greg Corrado, Macduff Hughes, and Jeffrey Dean. 2016.
\newblock \href {http://arxiv.org/abs/1609.08144} {Google's neural machine
  translation system: Bridging the gap between human and machine translation}.
\newblock \emph{CoRR}, abs/1609.08144.

\bibitem[{Zhang et~al.(2020)Zhang, Sun, Galley, Chen, Brockett, Gao, Gao, Liu,
  and Dolan}]{model-dialogpt}
Yizhe Zhang, Siqi Sun, Michel Galley, Yen-Chun Chen, Chris Brockett, Xiang Gao,
  Jianfeng Gao, Jingjing Liu, and Bill Dolan. 2020.
\newblock \href {https://doi.org/10.18653/v1/2020.acl-demos.30} {{DIALOGPT} :
  Large-scale generative pre-training for conversational response generation}.
\newblock In \emph{Proceedings of the 58th Annual Meeting of the Association
  for Computational Linguistics: System Demonstrations}, pages 270--278,
  Online. Association for Computational Linguistics.

\end{thebibliography}
